\documentclass[sigconf, nonacm]{acmart}

\usepackage{multirow}
\usepackage[normalem]{ulem}
\usepackage{enumitem}
\usepackage{balance}
\usepackage{graphicx}
\usepackage{subcaption}

\AtBeginDocument{%
  \providecommand\BibTeX{{%
    \normalfont B\kern-0.5em{\scshape i\kern-0.25em b}\kern-0.8em\TeX}}}

\copyrightyear{2024}
\acmYear{2024}
\acmDOI{XXXXXXX.XXXXXXX}

\acmConference[arXiv]{arXiv preprint}{Feb 09,
  2024}{Shanghai}
\acmISBN{978-1-4503-XXXX-X/18/06}

\begin{document}

\title{Prompt Learning on Temporal Interaction Graphs}

\author{Xi Chen}
\authornote{Both authors contributed equally to this research.}
\email{x_chen21@m.fudan.edu.cn}
\orcid{0009-0003-6156-4811}
\author{Siwei Zhang}
\authornotemark[1]
\orcid{0009-0008-3994-083X}
\email{swzhang22@m.fudan.edu.cn}
\affiliation{%
  \institution{Shanghai Key Laboratory of Data Science , School of Computer Science, Fudan University}
  \streetaddress{}
  \city{Shanghai}
  \country{China}
}

\author{Yun Xiong}
\authornote{Corresponding author.}
\email{yunx@fudan.edu.cn}
\orcid{0000-0002-8575-5415}
\author{Xixi Wu}
\orcid{0000-0002-9935-5957}
\email{21210240043@m.fudan.edu.cn}
\affiliation{%
  \institution{Shanghai Key Laboratory of Data Science , School of Computer Science, Fudan University}
  \streetaddress{}
  \city{Shanghai}
  \country{China}
}

\author{Jiawei Zhang}
\email{jiawei@ifmlab.org}
\orcid{0000-0002-2111-7617}
\affiliation{%
  \institution{IFM Lab, \\ Department of Computer Science, University of California}
  \streetaddress{}
  \city{Davis}
  \state{CA}
  \country{USA}
}

\author{Xiangguo Sun}
\email{xgsun@se.cuhk.edu.hk}
\orcid{0000-0002-2224-4634}
\affiliation{%
  \institution{Department of Systems Engineering and Engineering Management, The Chinese University of Hong Kong}
  \streetaddress{}
  \city{Hong Kong SAR}
  \country{China}
}

\author{Yao Zhang}
\email{yaozhang@fudan.edu.cn}
\orcid{0000-0003-1481-8826}
\affiliation{%
  \institution{Shanghai Key Laboratory of Data Science , School of Computer Science, Fudan University}
  \streetaddress{}
  \city{Shanghai}
  \country{China}
}

\author{Feng Zhao}
\email{zhaofeng.zhf@antgroup.com}
\author{Yulin Kang}
\email{yulin.kyl@antgroup.com}
\affiliation{%
  \institution{Ant Group}
  \streetaddress{}
  \city{Shanghai/ Hangzhou}
  \country{China}
}

\renewcommand{\shortauthors}{Xi Chen and Siwei Zhang, et al.}
\newcommand{\thenmode}{pre-train, prompt}
\newcommand{\andmode}{pre-train, prompt-based fine-tune}
\newcommand{\Prompter}{TProG}
\newcommand{\ours}{TIGPrompt}

\begin{abstract}
Temporal Interaction Graphs (TIGs) are widely utilized to represent real-world systems, such as e-commerce and social networks. To facilitate representation learning on TIGs, researchers have proposed a series of TIG models.
However, these models are still facing two tough gaps between the pre-training and downstream predictions in their ``pre-train, predict'' training paradigm.
First, there is a temporal gap that exhibits limited accommodation ability to their timely predictions. This shortcoming severely undermines their applicability in distant future predictions on the dynamically evolving TIG data. Second, there is a semantic gap due to the lack of versatility in these pre-trained models to effectively cater to diverse downstream tasks. This hinders their practical applications, as they struggle to align with their learning and prediction capabilities across various application scenarios.

Recently, the ``pre-train, prompt'' paradigm has emerged as a lightweight mechanism for model generalization.
Therefore, applying this paradigm within TIGs is a potential solution to solve the aforementioned challenges.
However, the adaptation of this paradigm to TIGs is not straightforward. The prevalent application of prompting in static graph contexts falls short in temporal settings due to a lack of consideration for time-sensitive dynamics and a deficiency in expressive power. To address this issue, we introduce Temporal Interaction Graph Prompting (\textbf{TIGPrompt}), a versatile framework that seamlessly integrates with existing TIG models, bridging both the temporal and semantic gaps mentioned above.
In detail, we propose a temporal prompt generator to offer temporally-aware prompts for different tasks. These prompts stand out for their minimalistic design, relying solely on the fine-tuning of the prompt generator with very little supervision data, which is extremely efficient. To cater to varying computational resource demands, we propose an extended ``pre-train, prompt-based fine-tune'' paradigm, offering greater flexibility. 
Through extensive experiments involving multiple benchmarks, representative TIG models, and downstream tasks, 
our TIGPrompt demonstrates the SOTA performance and remarkable efficiency advantages.

\end{abstract}






\maketitle

\section{Introduction}

\begin{figure}[!t]
    \centering
    \begin{subfigure}[b]{1\linewidth}
        \includegraphics[width=8cm]{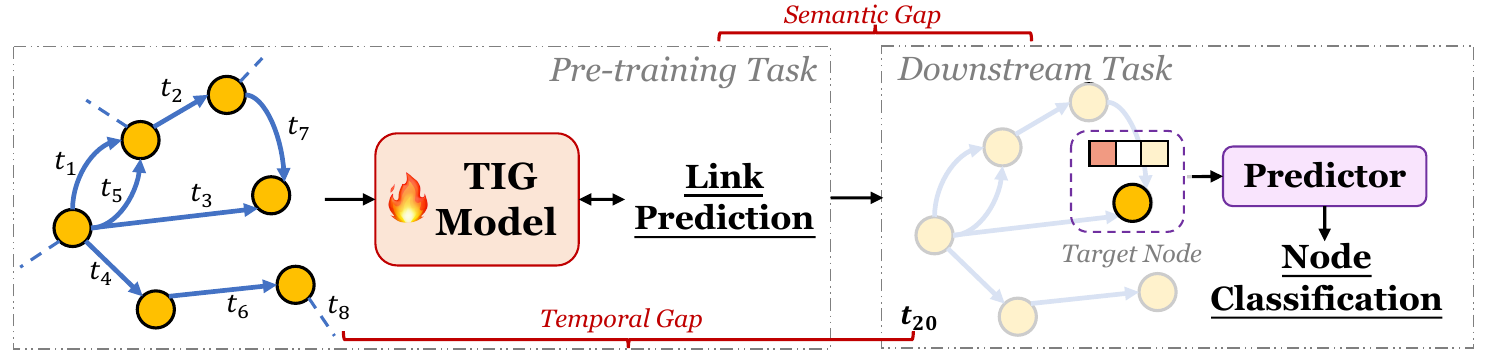}
        \caption{Existing ``pre-train, predict'' paradigm.}
    \end{subfigure}
    
    \vspace{0pt}
    \begin{subfigure}[b]{1\linewidth}
        \includegraphics[width=8cm]{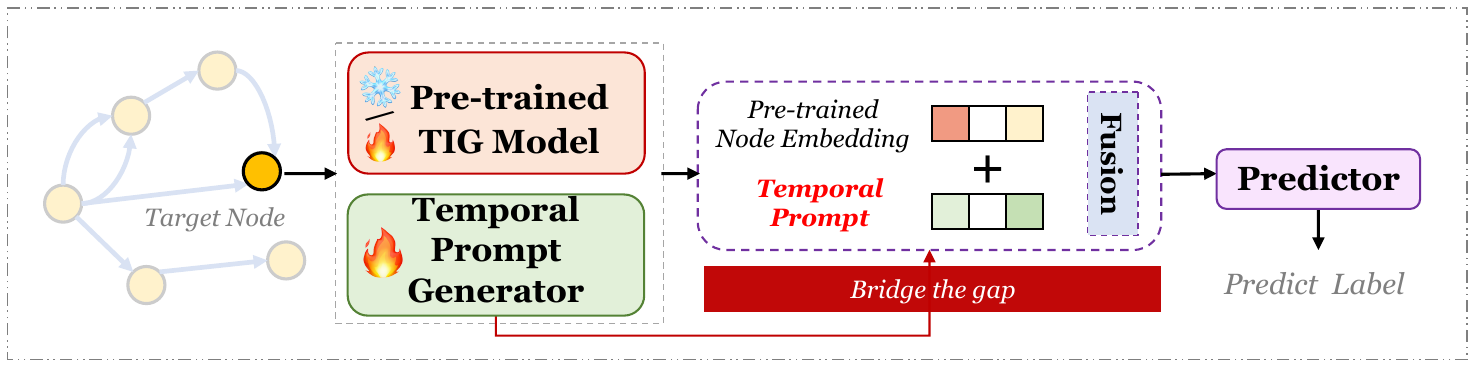}
        \caption{Our introduced prompting mechanism.}
    \end{subfigure}

\caption{(a) The ``pre-train, predict'' paradigm adopted by existing TIG models, which exhibits both temporal and semantic gaps when applied on the downstream task. 
(b) Our introduced prompting mechanism, with an innovative Temporal Prompt Generator, designed to mitigate both gaps.
}
\label{fig:overview}
\end{figure}

In real-world scenarios, interaction data is often accompanied by temporal information, i.e., timestamps, necessitating its modeling as Temporal Interaction Graphs (TIGs) \cite{ige, deepco}. In such a context, static graphs can hardly model such TIGs since they lack the necessary expressiveness to capture temporal dependencies. Specifically, in TIGs, objects are depicted as nodes, while timestamped interactions between these objects are represented as edges. Consequently, significant research efforts have been dedicated to TIG models \cite{jodie, dyrep, tgat, tgn, tiger}. These works aim to capture the dynamic nature of TIGs and learn temporal node representations, which can be applied to various downstream tasks~\cite{jodie, tgn, tiger}.

Recently, researchers have tried to explore the design of TIG models, leading to various effective exquisite TIG model structures \cite{tiger, ilore, rdgsl}. However, existing TIG models still follow the ``pre-train, predict'' learning framework, and they universally overlook the potential limitations of their basic training process. 
Fig. \ref{fig:overview}(a) exemplifies the traditional ``pre-train, predict'' paradigm, where a TIG model is pre-trained on a specific task (e.g., Link Prediction) and its learned knowledge is then transferred to various downstream tasks by tuning a corresponding predictor model (e.g., MLP \cite{mlp}).

However, this traditional training paradigm exists two huge gaps between such pretext and downstream tasks. First, the pre-training process will become severely outdated with temporal interactions \cite{tgl, speed}, leading to ineffective predictions for the distant future (i.e., \textbf{temporal gap}). This problem may require exhaustive re-training processes to incorporate new data recursively into model updating \cite{jodie, dyrep, tgat, tgn}, resulting in a significant consumption of computational resources. Second, the diversity in predictions, specifically for node- and edge-level tasks, significantly affects the performance of transferability across various tasks (i.e., \textbf{semantic gap}). For example, most of existing TIG models are pre-trained via edge-level strategies \cite{tgn, tiger}, but downstream tasks may be at the node-level \cite{tgn}. Intuitively, an edge-level pre-training strategy usually intends to make node representations smooth on observed edges, but there are many cases that two connected node has totally different labels, which might cause severe negative transfer \cite{04sun2023all, tiger}. Such misalignment hinders the adaptive ability of TIG models, thus limiting the model’s effectiveness in handling various downstream tasks.

To this end, the ``pre-train, prompt'' paradigm emerges as a panacea. 
Unlike the ``pre-train, predict'' paradigm, which requires substantial computational resources to re-train large well-trained models \cite{gnn, tgn, tiger, devlin2018bert, liu2023pre}, prompt learning only requires the design and training of lightweight prompts \cite{04sun2023all, 03liu2023graphprompt, 06fang2022universal} to adapt downstream tasks to the given pretext, thereby allowing the large, pre-trained model to remain unchanged.
This indicates that prompt learning can efficiently adapt a pre-trained model to handle evolving data at a significantly lower cost compared to re-training the model \cite{liu2023pre}. Furthermore, prompt vectors enable the explicit incorporation of task-specific knowledge \cite{04sun2023all}, offering greater flexibility than traditional learning frameworks.

Unfortunately, the applications of ``pre-train, prompt'' on graph domains are very limited and currently we only found a few works 
\cite{sun2023graph, 01sun2022gppt, 03liu2023graphprompt, 06fang2022universal}
that study prompting on static graphs, which is far from sufficient to be applied to TIGs. The reasons are two-fold: First, existing works proposed for static graphs overlook the temporal nature of TIGs, failing to inject temporal information into prompts to adapt to the evolving characteristics of TIGs \cite{deepco, jodie}. 
Second, current studies employ over-simplified prompt vectors for all nodes \cite{03liu2023graphprompt, 06fang2022universal}, which may suffice for static graphs; but for TIGs, where node representations normally keep evolving and require personalized updates along with timestamps, it falls short. 
Consequently, such simplified prompt vectors lack the expressiveness to accurately capture the personalized changing patterns for each unique node in TIGs.
Based on the above discussions, we observe that there are at least two technique challenges that prevent us from using prompt learning to solve the aforementioned problems. The \textbf{first challenge} is how to learn expressive prompts with the minimal cost to overcome the temporal gap caused by emerging data. The \textbf{second challenge} is how to design flexible and temporal-aware prompts that can support various TIG models and break down the semantic gap within diverse downstream application scenarios.

In this paper, as shown in Fig. \ref{fig:overview}(b), we introduce a new architecture for the TIG model to make it more generalized by bridging both temporal and semantic gaps with a novel approach, namely Temporal Interaction Graph Prompting (\textbf{\ours}). We overcome the weakness of existing prompting methods on static graphs and advance the application of graph prompting to temporal interaction graphs. In particular, we design a component named Temporal Prompt Generator (\Prompter), intelligently generating a personalized temporal prompt for each node in a TIG. Its implementation explicitly incorporates temporal information, allowing the prompt vector to effectively address the issue of temporal variability across different timestamps, thereby achieving greater expressiveness. Additionally, to bridge the semantic gap between pretext and downstream tasks, the \Prompter~component is tuned with a specific downstream task, facilitating adaptability to concrete downstream scenarios. 
Notably \ours~is extremely lightweight, as it involves only tuning the prompt generator while keeping the parameters of the TIG model frozen. Furthermore, it is weak-supervision tolerant, as it only requires a small portion of data for prompt tuning.
We further extend the proposed ``pre-train, prompt'' paradigm to cater to varying computational resource demands by introducing a ``pre-train, prompt-based fine-tune'' solution. The key distinction between these two paradigms lies in whether the parameters of the pre-trained TIG models are tuned during the prompt tuning stage. This approach allows us to achieve greater flexibility in accommodating different computational requirements in real-world deployment.

We summarize our contributions as follows:
\begin{itemize}[leftmargin=*, topsep=2pt]
   \item  We study the prompting mechanism on temporal interaction graphs. To the best of our knowledge, this is the first attempt that explores prompting on TIGs.

   \item We propose a ``pre-train, prompt'' paradigm specifically tailored for TIGs, bridging both temporal and semantic gaps in the traditional training process. Meanwhile, our framework supports various prompt generators, allowing for the temporal-aware personalized prompt vector for TIG prompt learning.

   \item To enhance the flexibility and accommodate diverse computational resources, we extend the paradigm to a ``pre-train, prompt-based fine-tune'' solution. Both paradigms are compatible with existing TIG models.

  \item Through extensive experiments on 4 benchmarks using 5 representative TIG models across 2 downstream tasks, our framework demonstrates the SOTA performance and remarkable efficiency.
\end{itemize}

\section{Related Work}
\label{sec:related_work_prompting}
\textbf{Temporal Interaction Graph Models.} Temporal Interaction Graph representation learning models (TIG models) are specifically designed to learn dynamic representations of the nodes in TIGs. These models employ node representations to execute downstream tasks, including link prediction (by computing node similarity) and node classification (through additional training of a classifier, i.e., projection head). The development of contemporary TIG models began with Jodie \cite{jodie}. 
Jodie utilizes two RNNs to dynamically update node representations and employs a projection operator to estimate the embeddings of nodes that have not interacted for an extended period. 
DyRep \cite{dyrep} introduces a deep temporal point process model, employing a dual-time scale approach to effectively capture both association and communication dynamics. 
TGAT \cite{tgat} revolutionizes TIG models by incorporating an attention mechanism, wherein it substitutes the original position coding with time coding to effectively aggregate information from a node's neighbors. 
Building on this, TGN \cite{tgn} introduces a memory module to store nodes' historical interaction information, and integrating these developments into a cohesive framework. 
TIGER \cite{tiger} presents a model equipped with a dual-memory module, specifically designed for enhanced aggregation of neighbor information. TIGER also introduces a restarter module, responsible for generating surrogate representations, which serve as a warm initialization for node representations. 
Additionally, several works are devoted to addressing challenges and resolving specific complexities inherent in TIG models, including large-scale training \cite{tgl, speed}, noise dynamics \cite{rdgsl}, and node-wise long-term modeling \cite{ilore} issues.
However, two critical issues persist: the limited adaptability of these models to new data, and the semantic gap between pretext tasks and downstream tasks.

\begin{figure*}[h]
\centering
\includegraphics[width=1\linewidth]{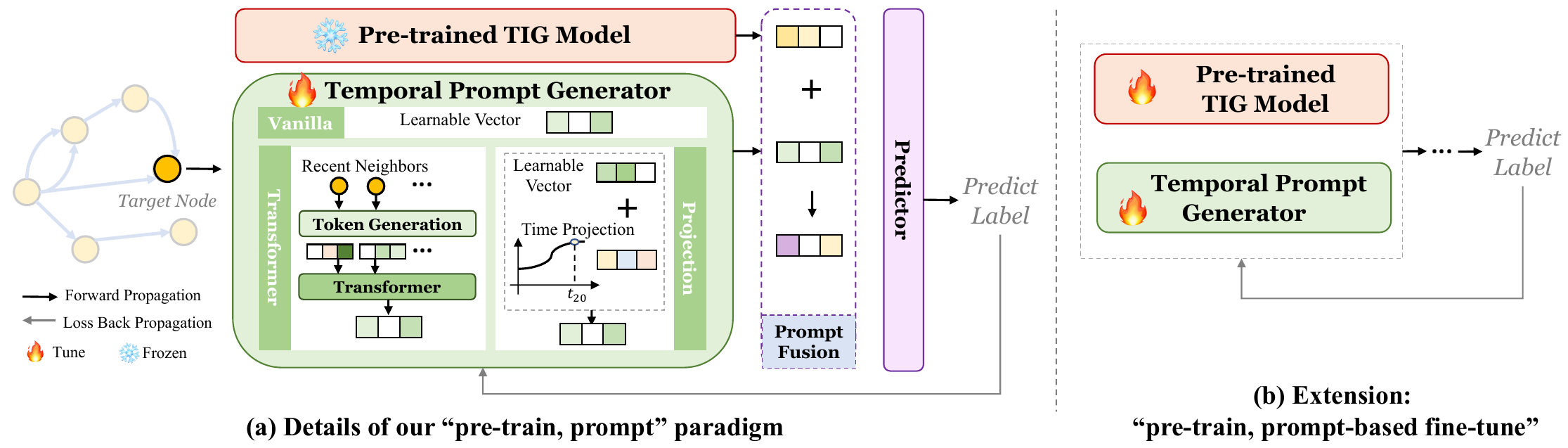}
\caption{Overview of TIGPrompt: (a) During the prompt tuning stage, the node embedding, calculated by the pre-trained TIG model, is combined with the personalized prompt embedding for downstream tasks. The TProG is optimized during this stage. (b) The key distinction between the two modes lies in whether the parameters of the TIG model are tuned.}
\label{fig:tigprompt}
\end{figure*}

\textbf{Graph Prompt Learning.} Prompt-tuning methods, originating from the NLP domain \cite{devlin2018bert, liu2023pre}, have gained widespread use in adapting pre-trained language models to a variety of downstream tasks. More recently, prompt learning has emerged in the graph domain \cite{qin2021learning, tsimpoukelli2021multimodal, 01sun2022gppt, 02zhu2023sgl, 03liu2023graphprompt, 04sun2023all, 05tan2023virtual, 06fang2022universal, 07huang2023prodigy, 08shirkavand2023deep, 09gong2023prompt, 10chen2023ultra, 11ma2023hetgpt, 12ge2023enhancing, 13yu2023hgprompt} as a promising approach for directing downstream tasks.
Pioneering works like GPPT \cite{01sun2022gppt} focus on the node classification task, incorporating learnable prompts directly into graphs.
Similarly, GraphPrompt \cite{03liu2023graphprompt} introduces a uniform prompt design, specifically tailored to address both node- and graph-level downstream tasks.
All-in-One \cite{04sun2023all} expands graph prompt learning further by encompassing prompt tokens, structures, and insertion patterns, introducing a comprehensive, albeit complex, prompting framework.
Nevertheless, there is a noticeable absence of prompt tuning methods specifically designed for the temporal interaction graphs, as existing static graph prompting works lack a temporal consideration and exhibit weak expressiveness.

\section{Proposed Method}
\label{sec:methods}

In this section, we elaborate on the detailed designs within the TIGPrompt framework. We first provide an overview of the ``\thenmode'' paradigm. Then, we show the implementation and optimization of our Temporal Prompt Generator (\Prompter) component, which enables the adaptability of pre-trained models across diverse downstream tasks. Finally, we extend this paradigm to the ``\andmode'' mode, specifically devised to accommodate varying computing resource constraints. An overview of our method is illustrated in Fig. \ref{fig:tigprompt}.

\subsection{``Pre-train, Prompt'' Paradigm Overview}

Existing TIG models such as JODIE \cite{jodie}, DyRec \cite{dyrep}, TGN \cite{tgn}, and TIGER \cite{tiger} primarily employ Link Prediction as the pre-training objective, with differences in their concrete model implementation. For instance, TGN \cite{tgn} introduces a memory-based approach and integrates previous works into a cohesive framework, while TIGER \cite{tiger} puts forward a model that incorporates a dual-memory module for more effective information aggregation. Once a TIG model is well-trained, node embeddings can be retrieved for task-specific predictions, such as node classification. The predictions are made as follows:

\begin{equation}
    \hat{\mathbf{Y}} = p_{\Phi}(\mathbf{Z}), \; \text{where} \; \mathbf{Z}= f_{\Theta}(\mathcal{V}, \mathcal{E}).
\end{equation}

\noindent Here, $p_{\Phi}(\cdot)$ denotes the projection head of the downstream task, $\mathbf{Z}$ denotes the learned node representations obtained from an arbitrary TIG model $f_{\Theta}(\cdot)$, which takes a TIG $G(\mathcal{V}, \mathcal{E})$ as the input. However, it is important to note that directly utilizing pre-trained node embeddings for downstream tasks is unfeasible as it overlooks two critical gaps: \textit{the temporal gap} (i.e., the evolving nature of TIGs may render pre-trained node embeddings less expressiveness to the timely TIG data), and \textit{the semantic gap} (i.e., the distinctions between link-level pretext task and node-level downstream task).

To bridge these gaps and enable the adaptability of a pre-trained TIG model across various scenarios, we propose to utilize personalized and temporal-aware \textit{prompt} for each node. Combined with pre-trained node embeddings, these prompts can carry task-specific semantics to get adapted to different downstream tasks as follows:

\begin{equation}
   \label{eq:fusion}
    \hat{\mathbf{Y}} = p_{\Phi}(\widetilde{\mathbf{Z}}), \; \widetilde{\mathbf{Z}} = f_{\rho}(\mathbf{Z}, \mathbf{P}),
\end{equation}

\noindent where $\mathbf{P}$ denotes the prompt matrix produced by the Temporal Prompt Generator, $f_{\rho}(\cdot)$ represents the fusion function, and $\widetilde{\mathbf{Z}}$ denotes the final prompted node representations. The prompt generator is tuned with task-specific supervision, enabling the final synthesized node representations contain task-specific and temporal-aware knowledge. Notably, during this process, the pre-trained TIG model $f_{\Theta}(\cdot)$ remains frozen, making TIGPrompt lightweight to get adapted to concrete downstream scenarios. Then, we move to the description of how these prompts are generated and tuned.

\subsection{\Prompter: Temporal Prompt Generator}

In this subsection, we provide a detailed explanation of our implementation of Temporal Prompt Generator, which produces a prompt matrix $\mathbf{P} \in \mathbb{R}^{|\mathcal{V}| \times d}$. We initially introduce a \textit{Vanilla} mode, where a learnable vector is assigned to each node, enabling personalized prompts tailored for specific downstream scenarios. To enhance the temporal awareness of produced prompts, we extend this mode by introducing two additional approaches: the \textit{Transformer} mode and the \textit{Projection} mode. 

\subsubsection{Vanilla \Prompter} 
We first introduce the simplest version of \Prompter, aimed at providing personalization expressiveness for each node. In this approach, the prompt for node $v \in \mathcal{V}$ is implemented as a learnable vector $\mathbf{p}_v \in \mathbb{R}^d$, which is initialized as zero vector and can be tuned to incorporate task-specific knowledge. This implementation bears a resemblance to traditional prompting techniques utilized in static graphs \cite{06fang2022universal, 03liu2023graphprompt}. Despite its simplicity, this method offers an intuitive design, easy implementation, and low parameterization, requiring only $\mathcal{O}(|\mathcal{V}|)$ parameters, scaling linearly with the size of the temporal interaction graph.

\subsubsection{Transformer \Prompter}

To generate temporal-aware prompt, we consider encoding the most relevant temporal information for each node. For a target node $v$, its most recent interactions provide valuable insights into its temporal information, which can be leveraged to generate the temporal prompt $\mathbf{p}_v$. 

Therefore, at any timestamp $t$, we first retrieve the node's most recent neighbor set $\mathcal{N}_v^t = \{ u | u \in \mathcal{V}, (u, v, t_{uv}) \in \mathcal{E}  \; \text{and} \; t_{uv} \leq t \}$. To avoid an excessively large neighbor set, we impose a restriction on the size of $\mathcal{N}_v^t$, returning only the most recent $K$ interactions. Then, for each neighboring node $u \in \mathcal{N}_v^t$, we first create a temporal neighbor token as follows:

 \begin{equation}
     \mathbf{t}_u = \mathbf{z}_v \; || \; \mathbf{z}_u \; || \; \mathbf{p}_u \; || \; \mathbf{e}_{uv} \; || \; f_{\omega}(t-t_{uv}),
 \end{equation}
 \noindent where $\mathbf{z}_u, \mathbf{z}_v$ are pre-trained node embeddings, $\mathbf{p}_u$ corresponds to the position embedding of node $u$ within the neighbor set, $\mathbf{e}_{uv}$ denotes the edge feature of historical interaction $(u,v, t_{uv})$, $||$ denotes the concatenation operation, and $f_{\omega}(\cdot)$ denotes a time encoding function (we apply the same time encoding method used in \cite{tiger}). In this way, the neighboring token $\mathbf{t}_u$ incorporates both interactive and temporal knowledge, and we further leverage a Transformer \cite{attention} to encode those temporal neighboring tokens to generate temporal prompt $\mathbf{p}_v$ as follows:

 \begin{equation}
     \mathbf{p}_v = \text{Transformer}( \{ \mathbf{t}_u | u \in \mathcal{N}_v^t \} ).
 \end{equation}
\noindent This approach ensures that the generated prompt $\mathbf{p}_v$ captures expressive temporal and recent interactive knowledge, promising to enhance downstream predictions. The implementation of Transformer \Prompter~is extremely lightweight, as the number of tunable parameters within this component is $\mathcal{O}(d)$, scaling linearly with the embedding dimension.

\subsubsection{Projection \Prompter} 
In addition to encoding recent neighboring information, we can also generate a temporal-aware prompt by integrating personalized vectors and time encoding. Recall that in the Vanilla mode, we introduce a learnable vector $\mathbf{p}_v^{\text{Personal}} \in \mathbb{R}^d$ for each node to represent the prompt. To incorporate the temporal knowledge, we fuse this personalized vector with time encoding. Specifically, at timestamp $t$, the temporal information can be encoded as $\mathbf{p}_{v}^{\text{Temporal}} = f_{\omega}(t - t_{v'})$, where $t_{v'}$ represents the most recent interaction timestamp of node $v$, and $f_{\omega}(\cdot)$ is a time encoding function. Finally, the temporal prompt $\mathbf{p}_v$ is generated via integrating both sides of information as follows:

\begin{equation}
    \mathbf{p}_v = \text{MLP}(  \mathbf{p}_v^{\text{Personal}} \; || \; \mathbf{p}_v^{\text{Temporal}} ),
\end{equation}

\noindent where $\text{MLP}(\cdot)$ \cite{mlp} is introduced to combine two types of information. The \textit{Projection} mode can be seen as a middle ground between the Vanilla mode and the Transformer mode, as it utilizes a learnable prompt vector to represent interactive information and a temporal vector to mimic the temporal evolution. Like the Vanilla mode, the number of tunable parameters required for the Projection \Prompter~is $\mathcal{O}(|\mathcal{V}|)$, scaling linearly with the size of the graph.

\subsection{Prompt Tuning and Inference}
Recall in Equation \ref{eq:fusion}, a fusion function is introduced to combine pre-trained node embeddings $\mathbf{Z}$ and prompt matrix $\mathbf{P}$ to yield prompted node representations. Specifically, we implement $f_{\rho}(\cdot)$ via a MLP parameterized by $\rho$ as follows:

\begin{equation}
    \widetilde{\mathbf{Z}} = f_{\rho}(\mathbf{Z}, \mathbf{P}) = \text{MLP}_{\rho}(  \mathbf{Z} \; || \; \mathbf{P} ),
\end{equation}

\noindent where $\widetilde{\mathbf{Z}}$ can be regarded as prompted embeddings, incorporating temporal knowledge to adapt to specific downstream tasks. 

Take the downstream Link Prediction task as an example, suppose a TIG has edge set $\mathcal{E}$, which can be split into three disjoint sets as $\mathcal{E} = \mathcal{E}^{\text{pre-train}} \cup \mathcal{E}^{\text{prompt}} \cup \mathcal{E}^{\text{val/test}}$. Here, $\mathcal{E}^{\text{pre-train}}$ denotes the set of edges used for pre-training the TIG model $f_{\Theta}(\cdot)$, $\mathcal{E}^{\text{prompt}}$ represents the set used to tune the prompt generator, and $\mathcal{E}^{\text{val/test}}$ denotes the edges for validation or testing. Specifically, given $\mathcal{E}^{\text{prompt}}$, the \Prompter~is optimized using predictions and ground-truth labels:

\begin{equation}
    \mathcal{L}_{\text{prompt-tune}}(\Phi, \rho, \mathbf{P})  = \text{Cross-Entropy}( p_{\Phi}(f_{\rho}( \mathbf{Z}, \mathbf{P} )),  \mathbf{Y}^{\text{prompt}}),
\end{equation}

\noindent where $\mathbf{Y}^{\text{prompt}}$ denotes the ground-truth labels provided by $\mathcal{E}^{\text{prompt}}$, $p_{\Phi}(\cdot)$ denotes the projection head of the Link Prediction task. Notably, during the prompt tuning stage, the TIG model remains frozen, avoiding exhaustive re-training processes. The tuning data only constitutes a small portion, meaning that even a small number of samples can help improve the adaptation of the pre-trained TIG model to downstream predictions. Similarly, the downstream Node Classification task can provide a small number of samples to tune \Prompter~and generate meaningful $\mathbf{P}$. Once \Prompter~is well-tuned, downstream predictions can be made as $\hat{\mathbf{Y}} = p_{\Phi}(f_{\rho}(\mathbf{Z}, \mathbf{P}))$.

By leveraging task-specific supervision to tune \Prompter, the prompts can incorporate task-specific semantics. This tuning process helps bridge both semantic and temporal gaps, resulting in improved downstream predictions.

\textbf{Extension: ``Pre-train, Prompt-based Fine-tune'' Paradigm.}
To accommodate to diverse computational resource requirements, we also extend the current ``\thenmode'' paradigm to the ``\andmode'' paradigm. The main difference between these two modes lies in whether the parameters of TIG model $f_{\Theta}(\cdot)$ is tuned during the prompt tuning stage. Therefore, for this paradigm, given prompt samples, both the prompts and the TIG model are optimized concurrently as follows:

\begin{equation}
    \mathcal{L}_{\text{fine-tune}}(\Phi, \rho, \mathbf{P}, \Theta)  = \text{Cross-Entropy}( p_{\Phi}(f_{\rho, \Theta}( \mathbf{Z}, \mathbf{P} )),  \mathbf{Y}^{\text{prompt}}).
\end{equation}

By jointly optimizing the TIG model and the prompts, these two components reinforce each other, leading to improved adaptability in various scenarios.

\subsection{Connection to Existing Graph Prompting Approaches}
\label{sec:methods_compare_sg}

As mentioned in the related works section (Sec. \ref{sec:related_work_prompting}), various prompting methods have been developed for static graphs. Most of these methods are specifically designed for a range of downstream tasks unique to static graph contexts. Among these methods, GraphPrompt \cite{03liu2023graphprompt} and GPF \cite{06fang2022universal} stand out as representatives and amenable to adaptation for the TIG model. GraphPrompt \cite{03liu2023graphprompt} utilizes a prompt vector on the outputted embeddings of GNN models, whereas GPF \cite{06fang2022universal} employs a similar prompt vector on the input data features. Therefore, in Sec. \ref{sec:comparison_static} we transfer these ideas to the TIG model and conduct experiments to see the comparable performance with our temporal graph prompting approach.

\section{Experiments}
\label{sec:exps}

\subsection{Datasets and Baselines}
We apply the proposed \ours~on four public datasets, Wikipedia,  Reddit, MOOC and LastFM \cite{jodie}. Detailed statistics of these datasets are presented in Appendix \ref{sec:app_data} (Tab. \ref{tab:data}). 
Only Wikipedia, Reddit and MOOC are with dynamic labels indicating state changes of users.
For datasets missing node or edge features, we adopt the approach used in prior works \cite{tgn, tiger}, representing them with zero vectors. 

For baseline comparisons, we select Jodie \cite{jodie}, DyRep \cite{dyrep}, TGN \cite{tgn}, TIGE \cite{tiger}. Additionally, we include TIGER-T \cite{tiger} as a baseline, considering it is a variant of TIGE and potentially offers improved performance over the TIGE model.

\subsection{Experimental Settings}
Our implementation and hyper-parameter settings are consistent with those in previous works \cite{tgn, tiger}. More information is discussed in Appendix \ref{sec:app_settings}.
Typically, the chosen baseline models split interaction edges chronologically into 70\% for training, 15\% for validation, and 15\% for testing. However, as discussed in Sec. \ref{sec:methods}, our aim is to demonstrate our method's adeptness in adapting to emerging data. For a fair comparison, we utilize the same data portion for training and inference. Therefore, we use only 50\% of the data for pre-training and 20\% for prompt tuning or fine-tuning, with the remaining 30\% equally divided for validation and testing. In essence, we train our model with less data and leverage a smaller portion for prompt tuning or fine-tuning to achieve enhanced performance on downstream tasks compared to the baselines.

\subsection{``Pre-train, Prompt''}

\begin{table*}[!t]
\centering
\caption{Under the ``\thenmode'' paradigm, results for the link prediction task — encompassing both transductive and inductive settings — are presented using Average Precision (\%). For the dynamic node classification task, results are measured in terms of AUROC (\%). The best performance is highlighted in bold.}
\setlength{\tabcolsep}{0.55mm}{

\begin{tabular}{cccccc|cccc||ccc}
\toprule
                                                                      &                                   & \multicolumn{4}{c|}{Transductive Link Prediction}                                       & \multicolumn{4}{c||}{Inductive Link Prediction}                                                & \multicolumn{3}{c}{Node Classification}                            \\ \midrule
                                                                      & \multicolumn{1}{c|}{\Prompter}    & Wikipedia                & Reddit             & MOOC               & LastFM             & Wikipedia                & Reddit             & MOOC                     & LastFM             & Wikipedia          & Reddit                   & MOOC               \\ \midrule
\multicolumn{1}{c|}{\multirow{4}{*}{\rotatebox[origin=c]{90}{Jodie}}} & \multicolumn{1}{c|}{Baseline}     & { 94.62±0.5}          & { 97.11±0.3}    & { 76.50±1.8}    & { 68.77±3.0}    & { 93.11±0.4}          & { 94.36±1.1}    & { 77.83±2.1}          & { 82.55±1.9}    & { 86.27±2.2}    & { 58.48±2.6}          & { 65.39±1.1}    \\
\multicolumn{1}{c|}{}                                                 & \multicolumn{1}{c|}{Vanilla} & 94.10±0.4                & 97.65±0.0          & 74.47±0.9          & 74.15±1.0          & 91.43±0.3                & 93.07±0.4          & 72.23±1.4                & 79.29±1.4          & 86.79±2.1          & \textbf{69.22±0.4}       & 69.21±0.4          \\
\multicolumn{1}{c|}{}                                                 & \multicolumn{1}{c|}{Transformer}  & \textbf{96.50±0.1}       & 98.28±0.0          & \textbf{82.90±1.1} & 77.98±2.1          & \textbf{95.08±0.2}       & 95.68±0.1          & 79.81±1.2                & 85.66±0.9          & 80.91±6.7          & 63.80±2.2                & 70.67±1.1          \\
\multicolumn{1}{c|}{}                                                 & \multicolumn{1}{c|}{Projection}   & 96.44±0.3                & \textbf{98.99±0.0} & 82.47±0.9          & \textbf{89.39±0.7} & 94.75±0.5                & \textbf{97.43±0.1} & \textbf{79.89±1.2}       & \textbf{92.62±0.4} & \textbf{87.08±1.1} & 68.26±0.9                & \textbf{76.45±0.6} \\ \midrule
\multicolumn{1}{c|}{\multirow{4}{*}{\rotatebox[origin=c]{90}{DyRep}}} & \multicolumn{1}{c|}{Baseline}     & { 94.59±0.2}          & { 97.98±0.1}    & { 75.37±1.7}    & { 68.77±2.1}    & { 92.05±0.3}          & { 95.68±0.2}    & { 78.55±1.1}          & { 81.33±2.1}    & { 85.11±1.4}    & { 62.77±2.1}          & { 66.68±3.4}    \\
\multicolumn{1}{c|}{}                                                 & \multicolumn{1}{c|}{Vanilla} & 89.64±1.0                & 97.63±0.0          & 71.57±2.7          & 72.62±1.1          & 85.45±1.2                & 92.92±0.3          & 71.34±0.5                & 77.31±1.7          & 84.88±1.4          & \textbf{65.67±2.4}       & 68.38±0.9          \\
\multicolumn{1}{c|}{}                                                 & \multicolumn{1}{c|}{Transformer}  & 94.51±0.4                & 98.27±0.0          & \textbf{80.59±1.9} & 76.89±1.6          & 92.44±0.4                & 95.73±0.1          & \textbf{78.89±0.2}       & 84.95±2.6          & 60.87±3.8          & 58.20±2.3                & 70.80±0.9          \\
\multicolumn{1}{c|}{}                                                 & \multicolumn{1}{c|}{Projection}   & \textbf{96.87±0.2}       & \textbf{99.06±0.0} & 79.76±1.9          & \textbf{89.04±0.6} & \textbf{95.37±0.3}       & \textbf{97.48±0.0} & 78.56±0.7                & \textbf{92.62±0.4} & \textbf{85.25±1.3} & 64.50±1.5                & \textbf{76.06±0.9} \\ \midrule
\multicolumn{1}{c|}{\multirow{4}{*}{\rotatebox[origin=c]{90}{TGN}}}   & \multicolumn{1}{c|}{Baseline}     & { \textbf{98.46±0.1}} & { 98.70±0.1}    & { 85.88±3.0}    & { 71.76±5.3}    & { \textbf{97.81±0.1}} & { 97.55±0.1}    & { \textbf{85.55±2.9}} & { 80.42±4.9}    & { 84.93±1.1}    & { 65.99±3.8}          & { 69.80±1.8}    \\
\multicolumn{1}{c|}{}                                                 & \multicolumn{1}{c|}{Vanilla} & 96.40±0.2                & 98.36±0.0          & 86.71±1.0          & 79.67±1.7          & 95.02±0.2                & 95.54±0.2          & 81.99±1.2                & 83.40±1.8          & 85.79±1.1          & \textbf{66.13±1.3}       & 70.16±1.9          \\
\multicolumn{1}{c|}{}                                                 & \multicolumn{1}{c|}{Transformer}  & 97.36±0.3                & 98.67±0.0          & 89.21±0.7          & 81.63±0.6          & 96.19±0.4                & 96.68±0.2          & 83.35±0.9                & 84.86±1.3          & 86.39±1.8          & 64.89±1.1                & 71.13±1.4          \\
\multicolumn{1}{c|}{}                                                 & \multicolumn{1}{c|}{Projection}   & 97.83±0.1                & \textbf{99.29±0.0} & \textbf{89.28±0.8} & \textbf{91.85±0.3} & 96.79±0.2                & \textbf{98.14±0.1} & 84.49±1.0                & \textbf{93.23±0.6} & \textbf{87.09±0.4} & 66.07±1.5                & \textbf{73.44±1.4} \\ \midrule
\multicolumn{1}{c|}{\multirow{4}{*}{\rotatebox[origin=c]{90}{TIGE}}}  & \multicolumn{1}{c|}{Baseline}     & { 98.83±0.1}          & { 99.04±0.0}    & { 89.64±0.9}    & { 87.85±0.9}    & { 98.45±0.1}          & { 98.39±0.1}    & { 89.51±0.7}          & { 90.14±1.0}    & { 83.98±3.4}    & { \textbf{65.36±2.9}} & { 69.61±2.5}    \\
\multicolumn{1}{c|}{}                                                 & \multicolumn{1}{c|}{Vanilla} & 98.75±0.0                & 98.88±0.0          & 88.91±0.4          & 89.54±0.3          & 98.22±0.0                & 97.73±0.0          & 88.22±0.3                & 90.64±0.2          & 86.18±0.5          & 62.13±2.0                & 70.57±1.1          \\
\multicolumn{1}{c|}{}                                                 & \multicolumn{1}{c|}{Transformer}  & 98.95±0.0                & 99.25±0.0          & \textbf{91.10±0.4} & 90.65±0.3          & 98.52±0.1                & 98.68±0.0          & 88.82±0.9                & 91.74±0.2          & 82.02±7.0          & 61.41±2.6                & 71.44±0.6          \\
\multicolumn{1}{c|}{}                                                 & \multicolumn{1}{c|}{Projection}   & \textbf{99.10±0.1}       & \textbf{99.47±0.0} & 90.94±0.2          & \textbf{95.21±0.2} & \textbf{98.75±0.1}       & \textbf{99.07±0.0} & \textbf{89.61±0.4}       & \textbf{95.82±0.1} & \textbf{86.65±0.9} & 60.75±1.3                & \textbf{75.18±2.1} \\ \midrule
\multicolumn{1}{c|}{\multirow{4}{*}{\rotatebox[origin=c]{90}{TIGER}}} & \multicolumn{1}{c|}{Baseline}     & { 98.90±0.0}          & { 99.02±0.0}    & { 86.99±1.6}    & { 85.17±0.2}    & { 98.58±0.0}          & { 98.59±0.0}    & { 86.42±1.7}          & { 89.11±0.3}    & { 80.84±4.6}    & { 62.58±1.3}          & { 64.91±5.2}    \\
\multicolumn{1}{c|}{}                                                 & \multicolumn{1}{c|}{Vanilla} & 98.89±0.0                & 98.90±0.0          & 87.43±0.4          & 86.13±0.4          & 98.50±0.0                & 98.33±0.0          & 87.28±1.5                & 88.45±0.4          & 85.12±0.3          & \textbf{63.16±1.4}       & 68.68±1.9          \\
\multicolumn{1}{c|}{}                                                 & \multicolumn{1}{c|}{Transformer}  & 98.98±0.0                & 99.22±0.0          & \textbf{90.31±0.4} & 88.22±0.4          & 98.59±0.0                & 98.88±0.0          & 89.05±1.0                & 90.66±0.3          & 77.15±8.9          & 61.94±2.1                & 71.26±1.2          \\
\multicolumn{1}{c|}{}                                                 & \multicolumn{1}{c|}{Projection}   & \textbf{99.16±0.0}       & \textbf{99.49±0.0} & 89.74±0.5          & \textbf{93.73±0.2} & \textbf{98.89±0.0}       & \textbf{99.26±0.0} & \textbf{89.42±1.5}       & \textbf{95.07±0.2} & \textbf{86.30±0.8} & 62.75±1.5                & \textbf{74.07±0.5} \\ \bottomrule
\end{tabular}

}

\label{tab:thencomb}
\end{table*}
\textbf{Link Prediction.} In the initial set of experiments, we keep the established protocols \cite{tgn, tiger} to assess model performance in both transductive and inductive temporal link prediction tasks.
In the transductive setting, we focus on those edges linked to nodes previously encountered in the training dataset. Conversely, in the inductive setting, the predictions center on temporal links between nodes that are unseen during the training phase. The evaluation metric is the average precision (AP) score. 

Adhering to the proposed ``\thenmode'' training paradigm, we keep the pre-trained model's parameters frozen during prompt tuning phrase.
As illustrated in Tab. \ref{tab:thencomb},
the experiments utilize three distinct proposed \Prompter s, respectively. Notably, the integration of prompts generated by the proposed \Prompter s with the original node representations results in significant improvements in downstream tasks. This approach yields SOTA results across nearly all datasets and baselines. 
This effectiveness stems from the fact that the prompts generated by the proposed TProG comprehensively incorporate temporal information, thereby bridging the temporal gap between pre-training and downstream task data.
Particularly for the LastFM dataset, where performance is previously sub-optimal, our method enhances performance by 29\% compared to prior approaches, as evidenced on two baselines. 
The fact that only a small portion of the data is used for prompt tuning underscores the efficacy of our methods, particularly the Transformer \Prompter~ and Projection \Prompter, in facilitating model adaptation to evolving timely TIG data.
However, in certain specific dataset/model combinations, such as Wikipedia/TGN, our model does not surpass the baseline. This limitation arises because breaking down the temporal gap in these contexts adversely affects the results. However, in the node classification experiments discussed subsequently, both temporal and semantic gaps exist between the pretext and downstream tasks. In these cases, our model achieves superior performance, indicating that in such scenarios, the semantic gap predominates as the primary limiting factor for the performance of the TIG model.

\textbf{Node Classification.} Dynamic node classification is conducted aiming to predict dynamic labels of nodes. It is utilized as a downstream task to validate the prompt's effectiveness and to demonstrate how the proposed method effectively bridges the temporal and semantic gap between pretext and downstream tasks. 
We conduct dynamic node classification on datasets with dynamic labels, i.e., Wikipedia, Reddit, and MOOC.

We use the same pre-trained models as in the link prediction task. The \Prompter s, however, are exclusively initialized and trained during the node classification process.
Following the approach in \cite{tgat, tgn, tiger}, we pass time-aware representations through a two-layer MLP to determine the probabilities of dynamic labels. However, these time-aware representations are substituted with prompted node representations, generated by the \Prompter.
In the original methodology, validation, and testing phases are not distinct, with the last epoch's results under a fixed maximum number of epochs being directly used for testing. To incorporate \Prompter~ training into this process, we adapt the validation and testing phases to mirror the link prediction task approach, allocating 15\% of the data for validation and another 15\% for testing. Concurrently, the baseline settings align with those used here.

As the results shown in Tab. \ref{tab:thencomb}, 
our method significantly outperforms the baseline on almost all node classification tasks, achieving the SOTA performance. It is worth noting that on the Reddit dataset, the Vanilla \Prompter~ alone is sufficient to achieve superior results. At the same time, the Projection \Prompter~ not only surpasses the baseline on Reddit but also shows the best performance on the other two datasets. For the MOOC dataset, our method improves upon the DyRep baseline by 15\%. These results demonstrate the substantial impact of the proposed training paradigm in bridging the gap between pretext and downstream tasks.

\subsection{``Pre-train, Prompt-based Fine-tune''}

\begin{table*}[!t]
\centering
\caption{Average Precision (\%) for the link prediction task under the ``\andmode'' paradigm.}
\begin{tabular}{cccccc|cccc}
\toprule
                                                                      &                                  & \multicolumn{4}{c|}{Transductive }                                       & \multicolumn{4}{c}{Inductive }                                           \\ \midrule
                                                                      & \multicolumn{1}{c|}{\Prompter}   & Wikipedia                & Reddit             & MOOC               & LastFM             & Wikipedia                & Reddit             & MOOC               & LastFM             \\ \midrule
\multicolumn{1}{c|}{\multirow{4}{*}{\rotatebox[origin=c]{90}{Jodie}}} & \multicolumn{1}{c|}{Baseline}    & { 94.62±0.5}          & { 97.11±0.3}    & { 76.50±1.8}    & { 68.77±3.0}    & { 93.11±0.4}          & { 94.36±1.1}    & { 77.83±2.1}    & { 82.55±1.9}    \\
\multicolumn{1}{c|}{}                                                 & \multicolumn{1}{c|}{Vanilla}     & 94.22±0.9                & 97.17±0.3          & 76.32±1.6          & 74.45±1.3          & 92.66±1.0                & 93.91±0.9          & 74.58±2.5          & 81.27±1.0          \\
\multicolumn{1}{c|}{}                                                 & \multicolumn{1}{c|}{Transformer} & \textbf{97.01±0.4}       & 98.25±0.1          & \textbf{85.52±0.6} & 76.48±1.5          & \textbf{96.13±0.5}       & 96.71±0.3          & \textbf{84.33±0.6} & 84.63±1.3          \\
\multicolumn{1}{c|}{}                                                 & \multicolumn{1}{c|}{Projection}  & 96.72±0.6                & \textbf{98.84±0.1} & 83.03±0.3          & \textbf{88.82±0.5} & 95.36±0.6                & \textbf{97.79±0.2} & 81.72±1.3          & \textbf{92.51±0.4} \\ \midrule
\multicolumn{1}{c|}{\multirow{4}{*}{\rotatebox[origin=c]{90}{DyRep}}} & \multicolumn{1}{c|}{Baseline}    & { 94.59±0.2}          & { 97.98±0.1}    & { 75.37±1.7}    & { 68.77±2.1}    & { 92.05±0.3}          & { 95.68±0.2}    & { 78.55±1.1}    & { 81.33±2.1}    \\
\multicolumn{1}{c|}{}                                                 & \multicolumn{1}{c|}{Vanilla}     & 90.48±1.1                & 97.15±0.2          & 74.88±2.5          & 72.96±0.5          & 88.50±1.3                & 93.31±0.7          & 73.42±2.7          & 80.79±1.8          \\
\multicolumn{1}{c|}{}                                                 & \multicolumn{1}{c|}{Transformer} & 95.62±0.4                & 98.17±0.1          & \textbf{84.81±1.1} & 74.22±1.8          & 94.52±0.6                & 96.61±0.2          & \textbf{83.38±0.7} & 83.74±2.5          \\
\multicolumn{1}{c|}{}                                                 & \multicolumn{1}{c|}{Projection}  & \textbf{97.19±0.2}       & \textbf{98.96±0.1} & 82.53±1.7          & \textbf{88.83±0.4} & \textbf{96.11±0.3}       & \textbf{97.78±0.2} & 81.51±1.0          & \textbf{92.59±0.4} \\ \midrule
\multicolumn{1}{c|}{\multirow{4}{*}{\rotatebox[origin=c]{90}{TGN}}}   & \multicolumn{1}{c|}{Baseline}    & { \textbf{98.46±0.1}} & { 98.70±0.1}    & { 85.88±3.0}    & { 71.76±5.3}    & { \textbf{97.81±0.1}} & { 97.55±0.1}    & { 85.55±2.9}    & { 80.42±4.9}    \\
\multicolumn{1}{c|}{}                                                 & \multicolumn{1}{c|}{Vanilla}     & 97.72±0.2                & 98.32±0.1          & 88.58±1.1          & 72.69±5.0          & 96.94±0.1                & 96.51±0.3          & 87.89±0.9          & 78.97±3.9          \\
\multicolumn{1}{c|}{}                                                 & \multicolumn{1}{c|}{Transformer} & 98.25±0.1                & 98.68±0.1          & 89.95±1.7          & 77.79±3.2          & 97.59±0.2                & 97.62±0.1          & 89.11±1.2          & 83.48±2.4          \\
\multicolumn{1}{c|}{}                                                 & \multicolumn{1}{c|}{Projection}  & 98.38±0.1                & \textbf{99.29±0.0} & \textbf{90.00±1.4} & \textbf{90.08±0.9} & \textbf{97.81±0.1}       & \textbf{98.61±0.1} & \textbf{89.15±1.6} & \textbf{92.64±0.9} \\ \midrule
\multicolumn{1}{c|}{\multirow{4}{*}{\rotatebox[origin=c]{90}{TIGE}}}  & \multicolumn{1}{c|}{Baseline}    & { 98.83±0.1}          & { 99.04±0.0}    & { 89.64±0.9}    & { 87.85±0.9}    & { 98.45±0.1}          & { 98.39±0.1}    & { 89.51±0.7}    & { 90.14±1.0}    \\
\multicolumn{1}{c|}{}                                                 & \multicolumn{1}{c|}{Vanilla}     & 98.84±0.0                & 98.87±0.0          & 90.18±0.7          & 89.06±0.5          & 98.37±0.0                & 97.82±0.2          & 89.59±0.5          & 91.06±0.4          \\
\multicolumn{1}{c|}{}                                                 & \multicolumn{1}{c|}{Transformer} & 98.99±0.0                & 99.20±0.0          & \textbf{92.14±0.9} & 91.22±0.3          & 98.58±0.0                & 98.70±0.1          & \textbf{91.22±0.8} & 92.81±0.3          \\
\multicolumn{1}{c|}{}                                                 & \multicolumn{1}{c|}{Projection}  & \textbf{99.12±0.0}       & \textbf{99.48±0.0} & 91.68±0.4          & \textbf{95.30±0.1} & \textbf{98.84±0.0}       & \textbf{99.16±0.0} & 91.16±0.4          & \textbf{96.20±0.1} \\ \midrule
\multicolumn{1}{c|}{\multirow{4}{*}{\rotatebox[origin=c]{90}{TIGER}}} & \multicolumn{1}{c|}{Baseline}    & { 98.90±0.0}          & { 99.02±0.0}    & { 86.99±1.6}    & { 85.17±0.2}    & { 98.58±0.0}          & { 98.59±0.0}    & { 86.42±1.7}    & { 89.11±0.3}    \\
\multicolumn{1}{c|}{}                                                 & \multicolumn{1}{c|}{Vanilla}     & 98.90±0.0                & 98.84±0.0          & 85.12±1.1          & 85.59±0.5          & 98.49±0.1                & 98.13±0.1          & 84.37±0.8          & 88.43±0.6          \\
\multicolumn{1}{c|}{}                                                 & \multicolumn{1}{c|}{Transformer} & 99.05±0.0                & 99.18±0.0          & 87.00±0.9          & 87.84±0.2          & 98.68±0.0                & 98.78±0.0          & 86.07±1.0          & 90.50±0.3          \\
\multicolumn{1}{c|}{}                                                 & \multicolumn{1}{c|}{Projection}  & \textbf{99.17±0.0}       & \textbf{99.49±0.0} & \textbf{87.83±0.6} & \textbf{93.50±0.2} & \textbf{98.88±0.0}       & \textbf{99.28±0.0} & \textbf{87.38±0.9} & \textbf{94.90±0.3} \\ \bottomrule
\end{tabular}
\label{tab:andlp}

\end{table*}
In the ``\andmode'' paradigm, we follow a similar experimental setting as with ``\thenmode'', with a key difference: instead of freezing the pre-trained model's parameters, we allow for their simultaneous optimization while using 20\% of the data to train the \Prompter. This adjustment aims to enhance the model's adaptability to new data and downstream tasks. 

As Tab. \ref{tab:andlp} shows, this paradigm yields improved results compared to ``\thenmode'', attributable to the fine-tuning of the pre-trained model. However, this approach requires more training resources due to the optimization of the pre-trained model's parameters. Thus, this training paradigm is recommended when sufficient resources are available to achieve optimal results. The experiments of the node classification task are discussed in the Appendix. \ref{sec:app_and_node}.

\subsection{Effectiveness of Various TProGs}
As indicated in Tab. \ref{tab:thencomb} and \ref{tab:andlp}, the Projection \Prompter~ generally outperforms other types of \Prompter~ in link prediction tasks, with the Transformer \Prompter~ also excelling in certain scenarios. In contrast, the Vanilla \Prompter~ often shows weaker performance, likely due to its limited capacity to express temporal information. 
However, in node classification tasks, the Vanilla \Prompter~ demonstrates improved results on specific datasets. Meanwhile, the Projection \Prompter~ consistently surpasses the baseline, though the Transformer \Prompter~ shows slightly lower effectiveness. 

The Transformer \Prompter~captures recent behavior patterns, whereas the Projection \Prompter~emphasizes the global historical state.
The scenarios where the Transformer \Prompter~demonstrates superior performance are predominantly observed on the MOOC dataset. This suggests that the recent behavioral characteristics inherent to this dataset are particularly effective in bridging the existing gaps.
The robust performance of the Projection \Prompter~ across various tasks can be ascribed to its ability to model global historical information, which possesses significant expressive power for capturing temporal dynamics in TIGs. Additionally, its node-specific learnable embeddings play a pivotal role in effectively bridging the semantic gap between pretext and downstream tasks.

\subsection{Comparison with Existing Graph Prompts}
\label{sec:comparison_static}
\begin{figure}[!tbp]
    

    \centering
    \begin{subfigure}[b]{.45\linewidth}
        \includegraphics[width=\linewidth]{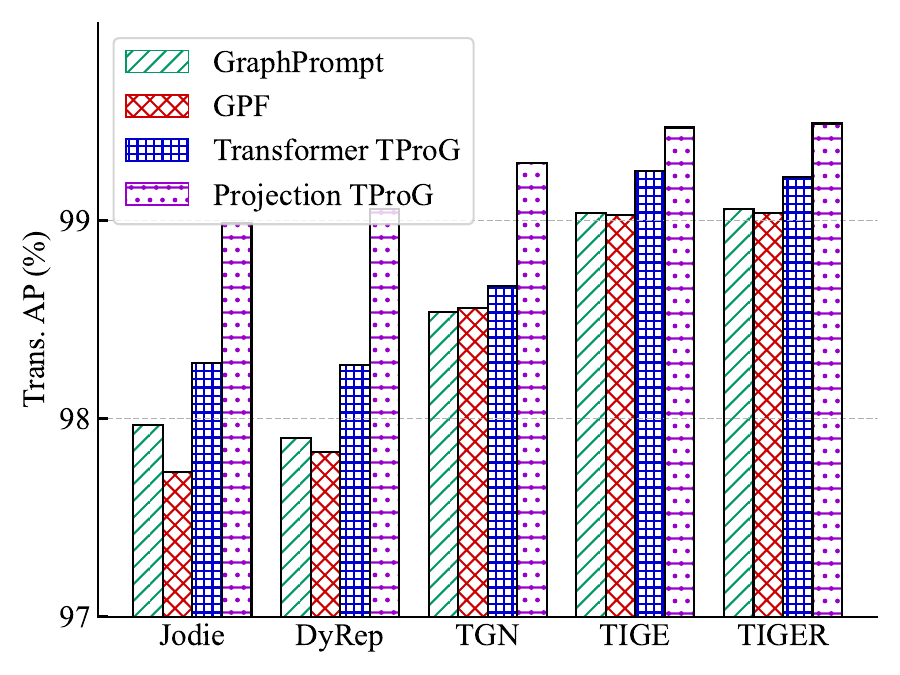}
        \caption{Reddit}
    \end{subfigure}
    \hspace{5pt}
    \begin{subfigure}[b]{.45\linewidth}
        \includegraphics[width=\linewidth]{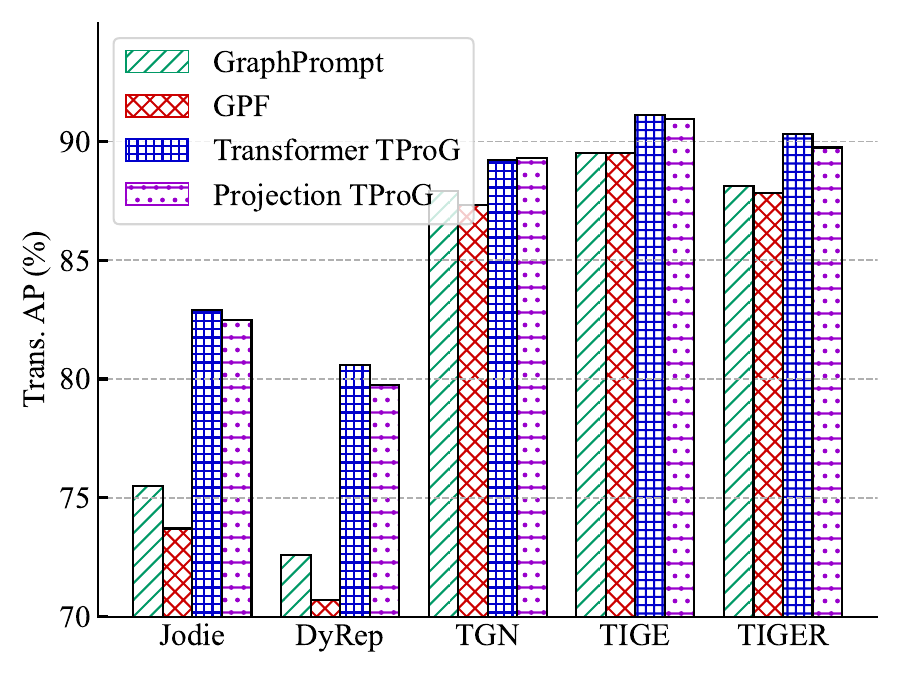}
        \caption{MOOC}
    \end{subfigure}

\caption{Comparison between traditional prompt on static graphs \cite{03liu2023graphprompt, 06fang2022universal} and our methods (``\thenmode'' paradigm, transductive link prediction on Reddit and MOOC datasets).}
\label{fig:para_sg}
\end{figure}
In addition, as discussed in Sec. \ref{sec:methods_compare_sg}, we conduct experiments using prompts from static graphs, i.e., GraphPrompt \cite{03liu2023graphprompt} and GPF \cite{06fang2022universal}, where a single, learnable prompt vector is applied uniformly across all nodes, either on the input \cite{06fang2022universal} or on the output \cite{03liu2023graphprompt} embeddings. The comparative results of these experiments are depicted in Fig. \ref{fig:para_sg}. The results demonstrate that our method significantly outperforms the traditional prompt method used in static graphs, demonstrating our effectiveness once again.

\subsection{Performance with Limited Data}

\begin{table*}[!t]
\centering
\caption{The results for the link prediction task under the ``\thenmode'' paradigm, note that only 20\% of data is used. Results colored in {\color[HTML]{3166FF} blue} indicate that they even surpass the baseline achieved with 70\% of the data used for training.}
\begin{tabular}{cccccc|cccc}
\toprule
\multicolumn{2}{c}{Only 20\% of data used}                                                                & \multicolumn{4}{c|}{Transductive}                                                                                                                                             & \multicolumn{4}{c}{Inductive}                                                                                                                          \\ \midrule
                                                                       & \multicolumn{1}{c|}{\Prompter}   & Wikipedia                                 & Reddit                                    & MOOC                                      & LastFM                                    & Wikipedia                                 & Reddit                                    & MOOC               & LastFM                                    \\ \midrule
\multicolumn{1}{c|}{}                                                  & \multicolumn{1}{c|}{Baseline}    & { 79.28±4.2}                           & { 92.39±1.4}                           & { 55.73±2.2}                           & { 68.00±0.7}                           & { 79.30±4.8}                           & { 80.58±2.8}                           & { 58.51±2.6}    & { 80.96±1.3}                           \\
\multicolumn{1}{c|}{}                                                  & \multicolumn{1}{c|}{Vanilla}     & 89.17±0.4                                 & 96.39±0.1                                 & 63.10±0.2                                 & {\color[HTML]{3166FF} 72.57±1.0}          & 88.00±0.6                                 & 94.33±0.1                                 & 63.52±0.3          & 77.13±0.9                                 \\
\multicolumn{1}{c|}{}                                                  & \multicolumn{1}{c|}{Transformer} & 92.11±0.9                                 & {\color[HTML]{3166FF} 97.54±0.0}          & 72.98±0.3                                 & {\color[HTML]{3166FF} 77.99±0.6}          & 92.34±0.7                                 & {\color[HTML]{3166FF} 96.43±0.0}          & 73.25±0.3          & 81.63±0.8                                 \\
\multicolumn{1}{c|}{\multirow{-4}{*}{\rotatebox[origin=c]{90}{Jodie}}} & \multicolumn{1}{c|}{Projection}  & {\color[HTML]{3166FF} \textbf{95.64±0.3}} & {\color[HTML]{3166FF} \textbf{98.54±0.1}} & \textbf{76.23±0.3}                        & {\color[HTML]{3166FF} \textbf{89.21±0.1}} & {\color[HTML]{3166FF} \textbf{95.04±0.2}} & {\color[HTML]{3166FF} \textbf{97.71±0.1}} & \textbf{76.31±0.3} & {\color[HTML]{3166FF} \textbf{90.94±0.2}} \\ \midrule
\multicolumn{1}{c|}{}                                                  & \multicolumn{1}{c|}{Baseline}    & { 88.19±1.0}                           & { 96.82±0.3}                           & { 73.13±1.7}                           & { 67.38±1.1}                           & { 85.99±0.9}                           & { 92.01±0.8}                           & { 71.91±2.1}    & { 79.67±1.8}                           \\
\multicolumn{1}{c|}{}                                                  & \multicolumn{1}{c|}{Vanilla}     & 84.27±1.2                                 & 96.35±0.1                                 & 61.19±1.3                                 & {\color[HTML]{3166FF} 69.85±0.5}          & 83.93±0.9                                 & 93.82±0.3                                 & 61.42±1.5          & 75.50±0.2                                 \\
\multicolumn{1}{c|}{}                                                  & \multicolumn{1}{c|}{Transformer} & 91.68±0.4                                 & 97.40±0.1                                 & 72.44±1.0                                 & {\color[HTML]{3166FF} 74.78±0.4}          & 91.23±0.5                                 & {\color[HTML]{3166FF} 96.28±0.2}          & 72.75±1.0          & 80.07±0.2                                 \\
\multicolumn{1}{c|}{\multirow{-4}{*}{\rotatebox[origin=c]{90}{DyRep}}} & \multicolumn{1}{c|}{Projection}  & {\color[HTML]{3166FF} \textbf{95.74±0.2}} & {\color[HTML]{3166FF} \textbf{98.63±0.0}} & {\color[HTML]{3166FF} \textbf{76.40±0.2}} & {\color[HTML]{3166FF} \textbf{88.26±0.2}} & {\color[HTML]{3166FF} \textbf{95.40±0.2}} & {\color[HTML]{3166FF} \textbf{97.74±0.1}} & \textbf{76.40±0.3} & {\color[HTML]{3166FF} \textbf{90.51±0.1}} \\ \midrule
\multicolumn{1}{c|}{}                                                  & \multicolumn{1}{c|}{Baseline}    & { 96.34±0.2}                           & { 97.63±0.1}                           & { 56.54±0.5}                           & { 66.54±2.0}                           & { 95.86±0.3}                           & { 95.98±0.4}                           & { 61.11±0.9}    & { 75.09±2.8}                           \\
\multicolumn{1}{c|}{}                                                  & \multicolumn{1}{c|}{Vanilla}     & 95.59±0.1                                 & 97.63±0.1                                 & 74.30±1.2                                 & 64.36±2.0                                 & 95.27±0.2                                 & 96.32±0.2                                 & 74.58±1.1          & 67.92±1.5                                 \\
\multicolumn{1}{c|}{}                                                  & \multicolumn{1}{c|}{Transformer} & 96.23±0.1                                 & 98.09±0.0                                 & 75.15±0.8                                 & 67.65±3.0                                 & 95.79±0.1                                 & 97.35±0.1                                 & 75.25±0.7          & 70.43±3.4                                 \\
\multicolumn{1}{c|}{\multirow{-4}{*}{\rotatebox[origin=c]{90}{TGN}}}   & \multicolumn{1}{c|}{Projection}  & \textbf{96.93±0.2}                        & {\color[HTML]{3166FF} \textbf{98.95±0.0}} & \textbf{79.10±0.6}                        & {\color[HTML]{3166FF} \textbf{87.42±0.4}} & \textbf{96.58±0.3}                        & {\color[HTML]{3166FF} \textbf{98.38±0.1}} & \textbf{79.17±0.5} & {\color[HTML]{3166FF} \textbf{88.65±0.6}} \\ \midrule
\multicolumn{1}{c|}{}                                                  & \multicolumn{1}{c|}{Baseline}    & { 98.36±0.1}                           & { 98.71±0.1}                           & { 80.60±1.5}                           & { 84.73±0.7}                           & { 98.11±0.1}                           & { 98.46±0.1}                           & { 80.71±1.4}    & { 85.73±0.8}                           \\
\multicolumn{1}{c|}{}                                                  & \multicolumn{1}{c|}{Vanilla}     & 98.50±0.0                                 & 98.58±0.0                                 & 80.58±0.4                                 & 85.24±0.3                                 & 98.20±0.0                                 & 98.16±0.0                                 & 80.88±0.3          & 86.45±0.0                                 \\
\multicolumn{1}{c|}{}                                                  & \multicolumn{1}{c|}{Transformer} & {\color[HTML]{3166FF} \textbf{98.92±0.0}} & {\color[HTML]{3166FF} 99.08±0.0}          & 80.32±1.2                                 & 87.77±0.4                                 & {\color[HTML]{3166FF} \textbf{98.69±0.0}} & {\color[HTML]{3166FF} 98.90±0.0}          & 80.56±1.1          & 88.81±0.3                                 \\
\multicolumn{1}{c|}{\multirow{-4}{*}{\rotatebox[origin=c]{90}{TIGE}}}  & \multicolumn{1}{c|}{Projection}  & 98.82±0.0                                 & {\color[HTML]{3166FF} \textbf{99.32±0.0}} & \textbf{83.11±0.1}                        & {\color[HTML]{3166FF} \textbf{93.40±0.2}} & {\color[HTML]{3166FF} 98.63±0.0}          & {\color[HTML]{3166FF} \textbf{99.16±0.0}} & \textbf{83.30±0.1} & {\color[HTML]{3166FF} \textbf{93.94±0.1}} \\ \midrule
\multicolumn{1}{c|}{}                                                  & \multicolumn{1}{c|}{Baseline}    & { 98.32±0.1}                           & { 98.67±0.1}                           & { 80.31±0.6}                           & { 84.53±0.4}                           & { 98.10±0.1}                           & { 98.12±0.2}                           & { 78.07±0.5}    & { 88.54±0.5}                           \\
\multicolumn{1}{c|}{}                                                  & \multicolumn{1}{c|}{Vanilla}     & 98.50±0.0                                 & 98.62±0.0                                 & 80.47±0.3                                 & 84.66±0.1                                 & 98.22±0.0                                 & 98.31±0.0                                 & 80.88±0.3          & 86.05±0.3                                 \\
\multicolumn{1}{c|}{}                                                  & \multicolumn{1}{c|}{Transformer} & {\color[HTML]{3166FF} \textbf{98.93±0.0}} & {\color[HTML]{3166FF} 99.04±0.0}          & 80.66±0.9                                 & {\color[HTML]{3166FF} 88.18±0.2}          & {\color[HTML]{3166FF} \textbf{98.67±0.0}} & {\color[HTML]{3166FF} 98.85±0.0}          & 80.90±0.9          & {\color[HTML]{3166FF} 89.36±0.2}          \\
\multicolumn{1}{c|}{\multirow{-4}{*}{\rotatebox[origin=c]{90}{TIGER}}} & \multicolumn{1}{c|}{Projection}  & 98.77±0.0                                 & {\color[HTML]{3166FF} \textbf{99.35±0.0}} & \textbf{82.96±0.3}                        & {\color[HTML]{3166FF} \textbf{93.10±0.1}} & 98.57±0.0                                 & {\color[HTML]{3166FF} \textbf{99.23±0.0}} & \textbf{83.16±0.3} & {\color[HTML]{3166FF} \textbf{93.73±0.0}} \\ \bottomrule
\end{tabular}

\label{tab:20then}
\end{table*}
\subsubsection{Performance with Limited Training Data}
To validate the effectiveness of the proposed prompt method and demonstrate that it requires only a small dataset to achieve superior results, we strategically design an experiment using merely 10\% of the data for pre-training, followed by another 10\% for prompt tuning (``\thenmode''). As a baseline for comparison, we utilized the results reported in TIGE \cite{tiger}, which is trained on only 20\% of the data. The experimental outcomes, detailed in Tab. \ref{tab:20then}, clearly illustrate that our method, even with limited data for training and prompt tuning, can attain the best results among all the baselines. Remarkably, on certain dataset/model combinations, our results even surpass the baseline achieved with 70\% of the data used for training.

\subsubsection{Performance with Limited Prompt Data}
\begin{figure}[!tbp]
    \centering
    \begin{subfigure}[b]{.45\linewidth}
        \includegraphics[width=\linewidth]{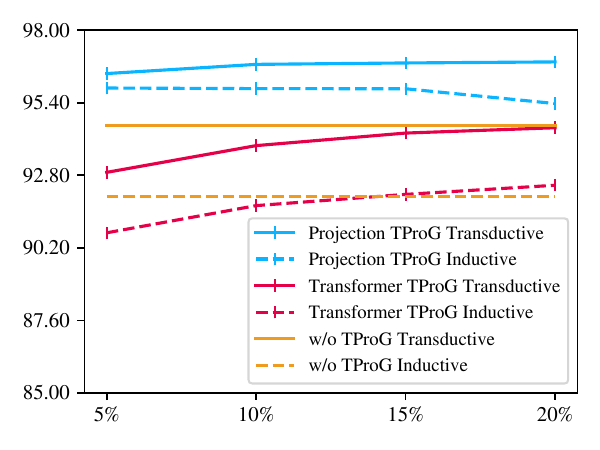}
        \caption{Wikipedia/DyRep}
    \end{subfigure}
    \hspace{5pt}
    \begin{subfigure}[b]{.45\linewidth}
        \includegraphics[width=\linewidth]{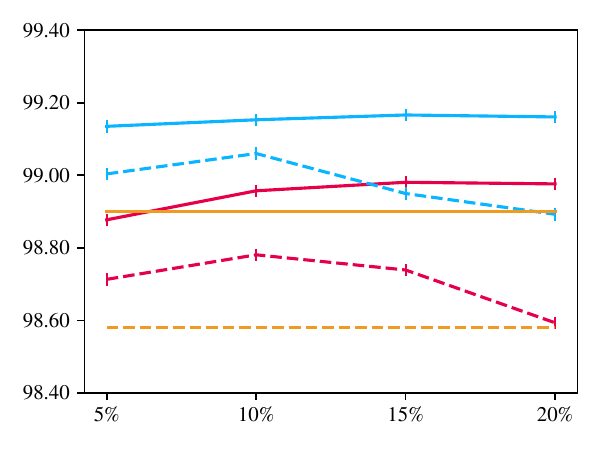}
        \caption{Wikipedia/TIGER}
    \end{subfigure}
    



    \caption{Performance w.r.t the Proportion of Prompting Data. This figure is continued in Appendix \ref{sec:app_para_analysis}.}
    \label{fig:para_data}
\end{figure}
To further explore the efficiency of our method, we investigate the minimum amount of data required for prompt tuning to surpass baseline performances. We utilize 50\% of the data for pre-training, and 5\% to 20\% data for prompt tuning. We select DyRep \cite{dyrep} and TIGER \cite{tiger} to conduct experiments under the ``\thenmode'' paradigm for this analysis. The results, as depicted in Fig. \ref{fig:para_data} and Fig. \ref{fig:para_data_app}, reveal that as little as 10\%, and in some cases only 5\%, of the data is needed for our approach to prompt tuning to achieve improved results. Furthermore, we observe that increasing the amount of data used for prompt tuning correspondingly enhances the performances in the transductive setting. This finding reaffirms the efficacy of our approach.

\subsection{Parameter Analysis}
\begin{figure}[!tbp]
    \centering
    \begin{subfigure}[b]{.45\linewidth}
        \includegraphics[width=\linewidth]{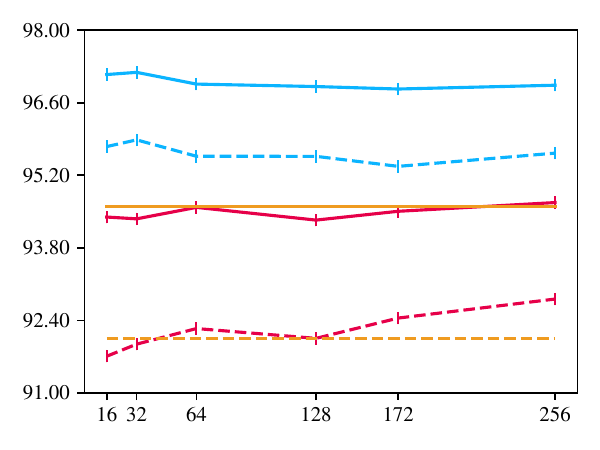}
        \caption{Wikipedia/DyRep}
    \end{subfigure}
    \hspace{5pt}
    \begin{subfigure}[b]{.45\linewidth}
        \includegraphics[width=\linewidth]{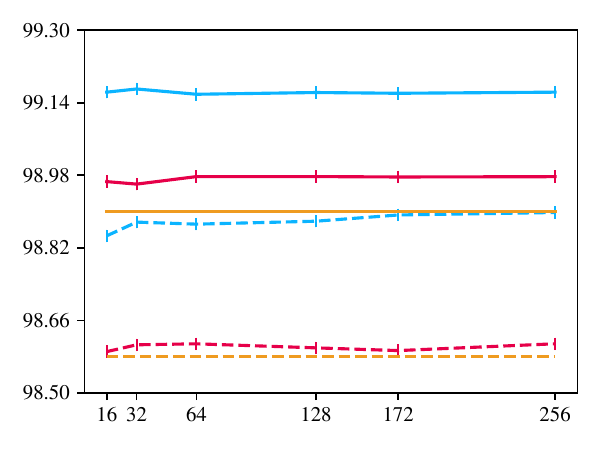}
        \caption{Wikipedia/TIGER}
    \end{subfigure}
    


\caption{Performance w.r.t the Prompts Dimension. This figure shares the same legend with Fig. \ref{fig:para_data} and is continued in Appendix \ref{sec:app_para_analysis}.}
\label{fig:para_dim}
\end{figure}
In these experiments, we explore the impacts of the dimension of the prompt vector. Additionally, we examine whether increasing the dimensions could yield even better results. As shown in Fig. \ref{fig:para_dim} and Fig. \ref{fig:para_dim_app}, the results indicate that a 64-dimensional prompt vector suffices to surpass the baseline performance in most cases. While higher dimensions do improve outcomes, they also increase the model's complexity. Researchers, therefore, should weigh the trade-off between experimental effectiveness and resource efficiency when selecting the optimal prompt vector dimension.

\subsection{Efficiency Analysis}

\begin{table}[t]
\centering
\caption{Training time for one epoch (in seconds) and trainable parameters count comparison.}
\begin{tabular}{cccc}
\toprule
                                                                            &                            TProG  & Training Time & Parameters Count   \\ \midrule
\multicolumn{1}{c|}{\multirow{4}{*}{{\rotatebox[origin=c]{90}{Wikipedia}}}} & \multicolumn{1}{c|}{Baseline}   & 63.4          & 451,1K           \\
\multicolumn{1}{c|}{}                                                       & \multicolumn{1}{c|}{Vanilla} & 4.6(-92.7\%)  & 326.4K(-27.7\%) \\
\multicolumn{1}{c|}{}                                                       & \multicolumn{1}{c|}{Transformer}  & 65.3(3.0\%)  & 338.4K(-25.0\%)  \\
\multicolumn{1}{c|}{}                                                       & \multicolumn{1}{c|}{Projection}   & 4.9(-92.3\%)  & 338.3K(-25.0\%)  \\ \midrule
\multicolumn{1}{c|}{\multirow{4}{*}{{\rotatebox[origin=c]{90}{MOOC}}}}      & \multicolumn{1}{c|}{Baseline}   & 138.4         & 454.3K           \\
\multicolumn{1}{c|}{}                                                       & \multicolumn{1}{c|}{Vanilla} & 12.0(-91.3\%) & 257.8K(-43.2\%)  \\
\multicolumn{1}{c|}{}                                                       & \multicolumn{1}{c|}{Transformer}  & 113.3(-18.1\%) & 341.4K(-24.8\%)  \\
\multicolumn{1}{c|}{}                                                       & \multicolumn{1}{c|}{Projection}   & 12.3(-91.1\%) & 269.7K(-40.6\%)  \\ \bottomrule
\end{tabular}

\label{tab:time}
\end{table}
We record the training time on the CPU and calculate the learnable parameters of the previous SOTA TIG model, TIGER \cite{tiger}, on two datasets.
As Tab. \ref{tab:time} shows, our method boosts efficiency and lowers training parameters versus the baselines. The Transformer \Prompter~exhibits modest time efficiency due to the inherent computational slowness of transformers. However, the other two \Prompter s both register substantial efficiency enhancements.
The results demonstrate that the proposed method is indeed lightweight. 

\section{Conclusion}

In this paper, we introduce two novel training paradigms for TIGs, which are grounded in pre-training, prompting, and fine-tuning techniques. Additionally, we present and compare three distinct temporal prompt generators, designed to ensure the resulting prompt vectors encapsulate a significant amount of temporal information. Employing the proposed paradigms can bridge both temporal and semantic gaps in the traditional training paradigm. Moreover, through extensive experimentation, we demonstrate that our methods significantly improve baseline results in TIG models across various downstream tasks, thus achieving the SOTA performance.
\balance




\balance{
\bibliographystyle{ACM-Reference-Format}
\bibliography{references}
}


\clearpage
\newpage

\appendix

\section{Datasets}
\label{sec:app_data}
\begin{table}[!h]
\centering
\caption{Dataset Statistic. $d_{n}$ and $d_{e}$ indicate the dim of nodes and edges, respectively.}
\begin{tabular}{c|ccccc}
\toprule
          & \# Nodes & \# Edges  & $d_{n}$ & $d_{e}$ & Classes \\ \midrule
Wikipedia & 9,227    & 157,474   & 172     & 172     & 2       \\
Reddit    & 10,984   & 672,447   & 172     & 172     & 2       \\
MOOC      & 7,144    & 411,749   & 172     & 172     & 2       \\
LastFM    & 1,980    & 1,293,103 & 172     & 172     & -       \\ \bottomrule
\end{tabular}
\label{tab:data}
\end{table}
In alignment with previous studies \cite{jodie, dyrep, tgn, tiger}, we utilize four public datasets made available by the authors of Jodie \cite{jodie}. Detailed statistics of these datasets can be found in Tab. \ref{tab:data}.

\section{Continued Experiment Results}
\label{sec:app_para_analysis}
We provide the complete experiment results for the limited prompt data analysis and parameter analysis here.

\begin{figure}[H]
    
    \begin{subfigure}[b]{.45\linewidth}
        \includegraphics[width=\linewidth]{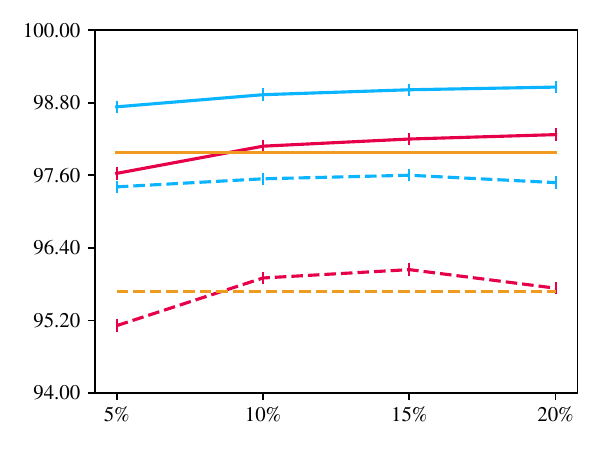}
        \caption{Reddit/DyRep}
    \end{subfigure}
    \hspace{5pt}
    \begin{subfigure}[b]{.45\linewidth}
        \includegraphics[width=\linewidth]{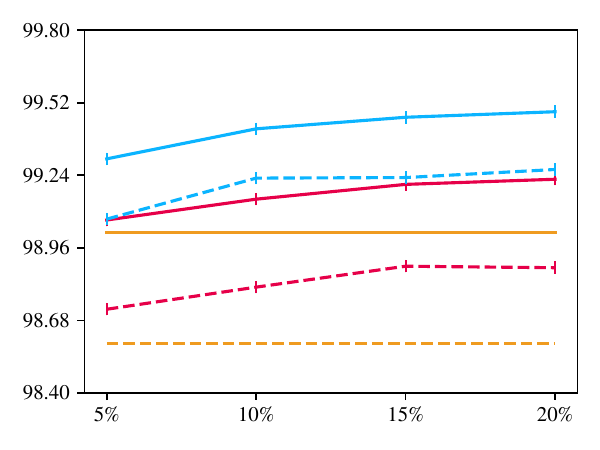}
        \caption{Reddit/TIGER}
    \end{subfigure}

    \vspace{0pt}
    \begin{subfigure}[b]{.45\linewidth}
        \includegraphics[width=\linewidth]{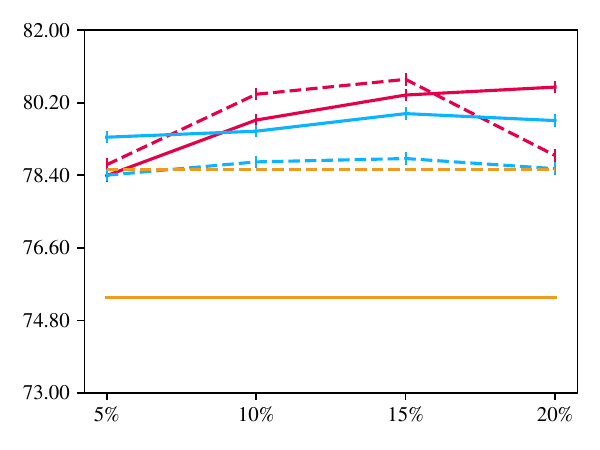}
        \caption{MOOC/DyRep}
    \end{subfigure}
    \hspace{5pt}
    \begin{subfigure}[b]{.45\linewidth}
        \includegraphics[width=\linewidth]{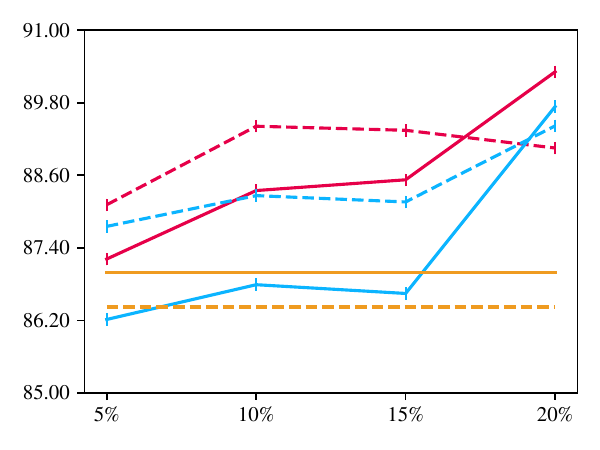}
        \caption{MOOC/TIGER}
    \end{subfigure}

    \vspace{0pt}
    \begin{subfigure}[b]{.45\linewidth}
        \includegraphics[width=\linewidth]{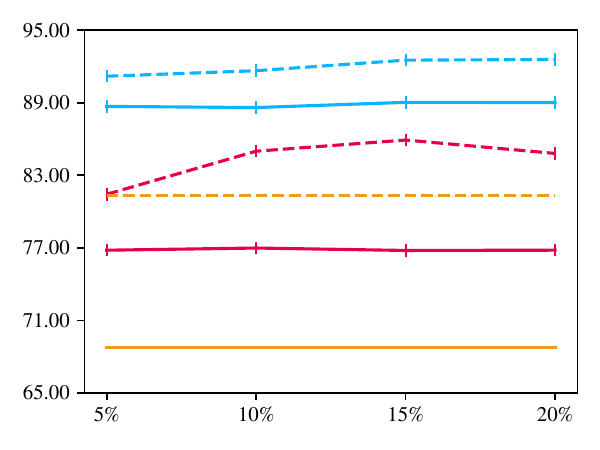}
        \caption{LastFM/DyRep}
    \end{subfigure}
    \hspace{5pt}
    \begin{subfigure}[b]{.45\linewidth}
        \includegraphics[width=\linewidth]{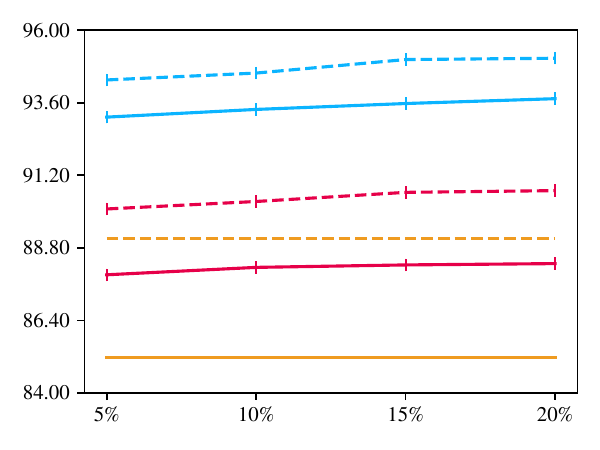}
        \caption{LastFM/TIGER}
    \end{subfigure}

    \caption{Performance w.r.t the Proportion of Prompting Data. This is a continued figure of Fig. \ref{fig:para_data}.}
    \label{fig:para_data_app}
\end{figure}
\begin{figure}[H]
    
    \begin{subfigure}[b]{.45\linewidth}
        \includegraphics[width=\linewidth]{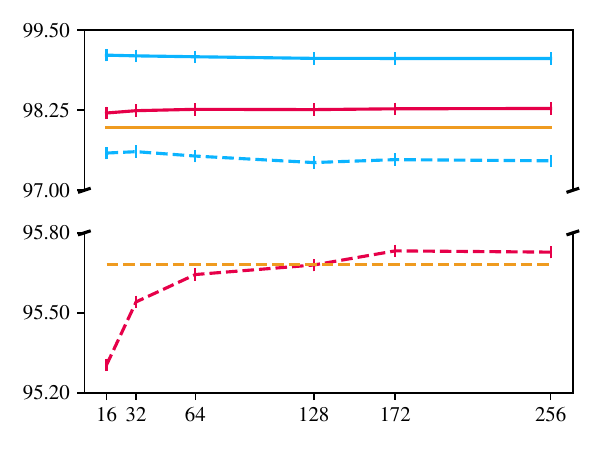}
        \caption{Reddit/DyRep}
    \end{subfigure}
    \hspace{5pt}
    \begin{subfigure}[b]{.45\linewidth}
        \includegraphics[width=\linewidth]{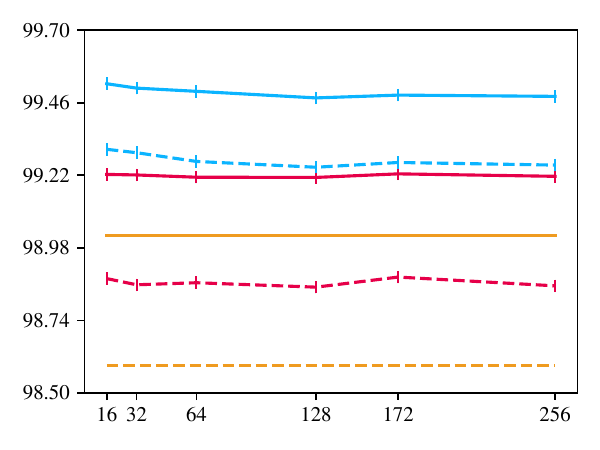}
        \caption{Reddit/TIGER}
    \end{subfigure}

    \vspace{0pt}
    \begin{subfigure}[b]{.45\linewidth}
        \includegraphics[width=\linewidth]{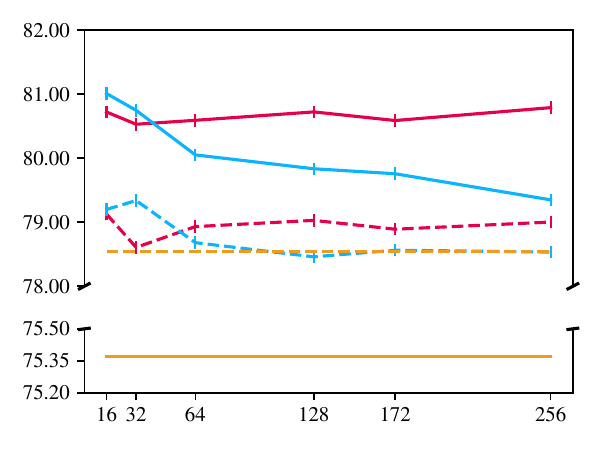}
        \caption{MOOC/DyRep}
    \end{subfigure}
    \hspace{5pt}
    \begin{subfigure}[b]{.45\linewidth}
        \includegraphics[width=\linewidth]{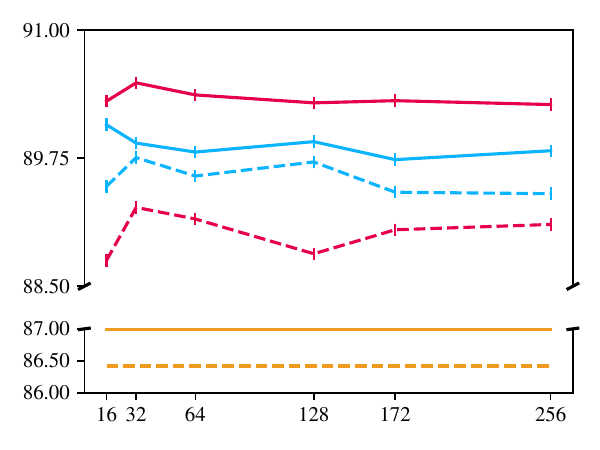}
        \caption{MOOC/TIGER}
    \end{subfigure}

    \vspace{0pt}
    \begin{subfigure}[b]{.45\linewidth}
        \includegraphics[width=\linewidth]{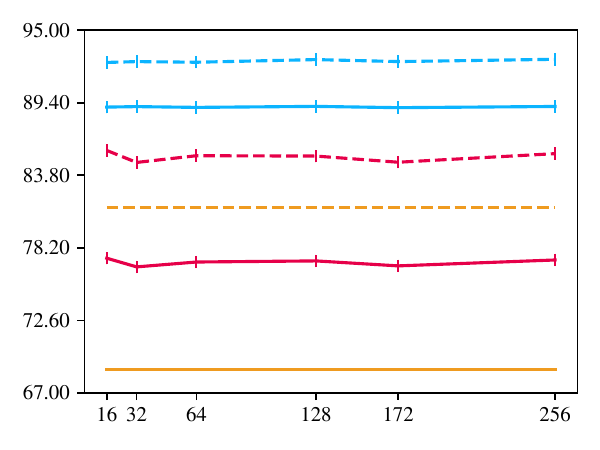}
        \caption{LastFM/DyRep}
    \end{subfigure}
    \hspace{5pt}
    \begin{subfigure}[b]{.45\linewidth}
        \includegraphics[width=\linewidth]{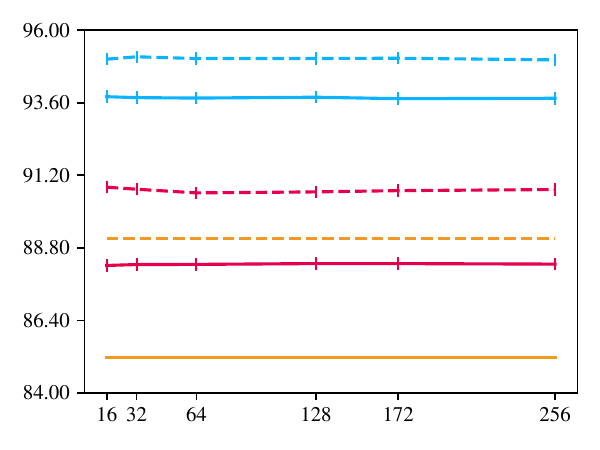}
        \caption{LastFM/TIGER}
    \end{subfigure}
\caption{Performance w.r.t the Prompts Dimension. This figure shares the same legend with Fig. \ref{fig:para_data}, and this is a continued figure of Fig. \ref{fig:para_dim}.}
\label{fig:para_dim_app}
\end{figure}

\section{Node Classification Under ``\andmode''}
\label{sec:app_and_node}

\subsection{Training Strategies}
\label{sec:app_and_node_strategies}
Under the ``\andmode'' paradigm for the node classification task, three different strategies can be applied: (1) directly employing the \Prompter~ trained in the link prediction task to generate prompts; (2) using the link prediction-trained \Prompter~ to initialize a \Prompter~ and then further optimizing it during node classification; and (3) discarding the previously \Prompter~ and re-initializing a new one for optimization alongside the node classification task. 


\begin{table}[t]
\centering
\caption{AUROC (\%) for dynamic node classification task under ``\andmode''. }
\begin{tabular}{cc|ccc}
\toprule
                                                                      & \Prompter    & Wikipedia                & Reddit                   & MOOC               \\ \midrule
\multicolumn{1}{c|}{\multirow{4}{*}{\rotatebox[origin=c]{90}{Jodie}}} & Baseline     & { 86.27±2.2}          & { 58.48±2.6}          & { 65.39±1.1}    \\
\multicolumn{1}{c|}{}                                                 & Vanilla/ Raw & 84.82±0.3                & 63.87±1.4                & 66.32±1.8          \\
\multicolumn{1}{c|}{}                                                 & Transformer  & \textbf{86.42±2.4}       & \textbf{67.19±1.0}       & 71.36±0.8          \\
\multicolumn{1}{c|}{}                                                 & Projection   & 84.41±3.0                & 62.27±3.8                & \textbf{75.89±1.5} \\ \midrule
\multicolumn{1}{c|}{\multirow{4}{*}{\rotatebox[origin=c]{90}{DyRep}}} & Baseline     & { 85.11±1.4}          & { 62.77±2.1}          & { 66.68±3.4}    \\
\multicolumn{1}{c|}{}                                                 & Vanilla/ Raw & \textbf{88.64±1.8}       & 58.64±2.7                & 65.00±2.2          \\
\multicolumn{1}{c|}{}                                                 & Transformer  & 83.73±0.3                & \textbf{64.58±2.2}       & 71.98±2.8          \\
\multicolumn{1}{c|}{}                                                 & Projection   & 85.35±0.5                & 58.84±2.1                & \textbf{75.09±1.3} \\ \midrule
\multicolumn{1}{c|}{\multirow{4}{*}{\rotatebox[origin=c]{90}{TGN}}}   & Baseline     & { \textbf{84.93±1.1}} & { \textbf{65.99±3.8}} & { 69.80±1.8}    \\
\multicolumn{1}{c|}{}                                                 & Vanilla/ Raw & 82.49±2.7                & 62.93±3.8                & 64.66±3.9          \\
\multicolumn{1}{c|}{}                                                 & Transformer  & 82.43±1.1                & 64.67±3.5                & 70.03±2.9          \\
\multicolumn{1}{c|}{}                                                 & Projection   & 83.86±1.4                & 60.28±4.8                & \textbf{77.15±3.1} \\ \midrule
\multicolumn{1}{c|}{\multirow{4}{*}{\rotatebox[origin=c]{90}{TIGE}}}  & Baseline     & { 83.98±3.4}          & { \textbf{65.36±2.9}} & { 69.61±2.5}    \\
\multicolumn{1}{c|}{}                                                 & Vanilla/ Raw & 81.43±6.8                & 62.46±2.5                & 70.35±0.8          \\
\multicolumn{1}{c|}{}                                                 & Transformer  & 85.87±2.0                & 64.14±1.6                & 67.61±5.9          \\
\multicolumn{1}{c|}{}                                                 & Projection   & \textbf{88.51±0.8}       & 59.08±3.9                & \textbf{78.04±3.2} \\ \midrule
\multicolumn{1}{c|}{\multirow{4}{*}{\rotatebox[origin=c]{90}{TIGER}}} & Baseline     & { 80.84±4.6}          & { 62.58±1.3}          & { 64.91±5.2}    \\
\multicolumn{1}{c|}{}                                                 & Vanilla/ Raw & 84.93±2.5                & \textbf{64.22±1.8}       & 68.16±2.9          \\
\multicolumn{1}{c|}{}                                                 & Transformer  & 83.95±4.4                & 60.75±1.3                & 68.26±1.8          \\
\multicolumn{1}{c|}{}                                                 & Projection   & \textbf{85.13±1.4}       & 61.20±2.2                & \textbf{81.58±1.2} \\ \bottomrule
\end{tabular}

\label{tab:andnode}
\end{table}

We choose the first strategy for our experiments, with the outcomes detailed in Tab. \ref{tab:andnode}. Notably, a part of these results exceed those achieved under the ``\thenmode'' paradigm. However, similar to the link prediction task, this approach demands additional training resources. A comparison of three training strategies is presented in Appendix \ref{sec:app_and_node_three_st}. This comparison demonstrates that applying the other two strategies has the potential to improve the performance of node classification tasks.

\subsection{Comparison Between Three Strategies of Node Classification Training}
\label{sec:app_and_node_three_st}

\begin{figure}[H]
\centering
\includegraphics[width=0.6\linewidth]{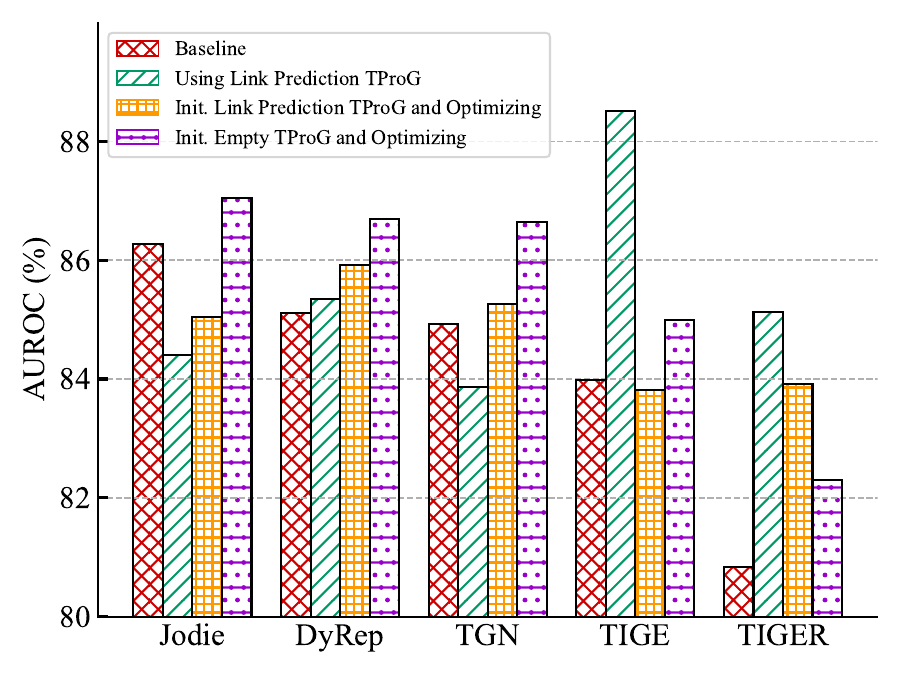}
\caption{Comparison between three different ``\andmode'' node classification training strategies (Wikipedia dataset, employing the Projection \Prompter).}
\label{fig:and_node_wiki_3}
\end{figure}

Beyond the initial experiments conducted under ``\andmode'' for the node classification task, we extend our investigation to include various training strategies outlined in Appendix \ref{sec:app_and_node_strategies}. A series of experiments was conducted using the Wikipedia dataset, employing the Projection \Prompter. The outcomes of these experiments are illustrated in Fig. \ref{fig:and_node_wiki_3}. The results indicate that our method outperforms the baseline models when different strategies are applied, thereby demonstrating the effectiveness of our approach.

\section{Implementation Details}
\label{sec:app_settings}
\begin{table}[h]
\centering
\caption{Default values of hyper-parameters.}
\begin{tabular}{c|c}
\toprule
Hyper-parameter           & Value  \\ \midrule
Batch size (Pre-training) & 200    \\
Batch size (Prompt tuning)  & 100    \\
Learning rate             & 0.0001 \\
Optimizer                 & Adam   \\
Prompt dimension          & 172    \\
Memory dimension          & 172    \\ \bottomrule
\end{tabular}

\label{tab:hyper}
\end{table}

We implement our methods using PyTorch, based on the official implementations of TGN \cite{tgn} and TIGER \cite{tiger}. Unless specified otherwise, we adhere to the default hyper-parameters listed in Tab. \ref{tab:hyper} for the experiments. For fair comparison, we maintain consistency in hyper-parameter settings with those used in TGN \cite{tgn} and TIGER \cite{tiger}.
Since we strictly follow the settings in TGN \cite{tgn} and TIGER \cite{tiger}, we reuse the results reported in \cite{tiger} as baselines.

All experiments are conducted on a single server with 72 cores, 32GB memory, and single Nvidia Tesla V100 GPU.

\end{document}